\documentclass[final,3p,times]{elsarticle}

\usepackage{tikz}
\usepackage{xifthen}
\usetikzlibrary{intersections}
\usepackage{tkz-euclide}

\pgfdeclarelayer{bg}    
\pgfsetlayers{bg,main}  
\definecolor{beige}{RGB}{245, 245, 220}
\definecolor{darkgrey}{RGB}{65, 65, 65}
\definecolor{lightgrey}{RGB}{250, 250, 250}
\newcommand*\circled[1]{\tikz[baseline=(char.base)]{
            \node[shape=circle,draw,inner sep=1pt, text width=3mm, align=center] (char) {#1};}}
\newcommand*\darkcircled[1]{\tikz[baseline=(char.base)]{
            \node[shape=circle,draw,inner sep=1pt, text width=3mm, align=center, fill=darkgrey, text=white, font=\bfseries] (char) {#1};}}

\newcommand*\bwcircled[1]{\tikz[baseline=(char.base)]{
            \node[shape=circle,double,draw,inner sep=1pt, text width=3mm, align=center] (char) {#1};}}

\usetikzlibrary{calc, arrows, fit, positioning, patterns, decorations.pathreplacing, shapes}
\tikzstyle{dash} = [dashed, -latex,>=latex]
\tikzstyle{line} = [draw, -latex,>=latex]
\tikzstyle{box} = [draw, minimum size=.8cm]
\tikzstyle{roundbox} = [draw, circle, inner sep=0pt, minimum size=3mm]
\tikzstyle{clamped} = [draw, fill=black, minimum size=0.15cm]
\tikzstyle{msgcircle} = [shape=circle, draw, inner sep=0pt, minimum size=5mm, fill=white]
\tikzstyle{darkmsgcircle} = [shape=circle, draw, inner sep=0pt, minimum size=5mm, fill=darkgrey, text=white, font=\bfseries]
\tikzstyle{msgdoublecircle} = [shape=circle, double, double distance=1.5pt, draw, inner sep=0pt, minimum size=5mm, fill=white]
\tikzstyle{darkmsgdoublecircle} = [shape=circle, double, double distance=1.5pt, draw, inner sep=0pt, minimum size=5mm, fill=darkgrey, text=white, font=\bfseries]

\newcommand{\parentheses}[4]{
      \draw[rounded corners=0.3cm, line width = 0.75pt] ($({#1},{#3})+(0,0.3)$) -- ($({#1},{#3})+(-0.3,0)$) -- ($({#1},{#4})+(-0.3,0)$) -- ($({#1},{#4})+(0,-0.3)$);
      \draw[rounded corners=0.3cm, line width = 0.75pt] ($({#2},{#3})+(0,0.3)$) -- ($({#2},{#3})+(0.3,0)$) -- ($({#2},{#4})+(0.3,0)$) -- ($({#2},{#4})+(0,-0.3)$);}

\newcommand{\msg}[6]{
      \ifthenelse{\isin{#1}{left} \AND \isin{#2}{down}}{
            \coordinate (anchor) at ($({#3})!{#5}!({#4})$);
            \node[msgcircle, xshift=-5.5mm] at (anchor) {#6};
            \node[xshift=-1.5mm] at (anchor) {$\downarrow$};
      }{}
      \ifthenelse{\isin{#1}{right} \AND \isin{#2}{down}}{
            \coordinate (anchor) at ($({#3})!{#5}!({#4})$);
            \node[msgcircle, xshift=5.5mm] at (anchor) {#6};
            \node[xshift=1.5mm] at (anchor) {$\downarrow$};
      }{}

      \ifthenelse{\isin{#1}{down} \AND \isin{#2}{right}}{
            \coordinate (anchor) at ($({#3})!{#5}!({#4})$);
            \node[msgcircle, yshift=-6.0mm] at (anchor) {#6};
            \node[yshift=-2.0mm] at (anchor) {$\rightarrow$};
      }{}
      \ifthenelse{\isin{#1}{up} \AND \isin{#2}{right}}{
            \coordinate (anchor) at ($({#3})!{#5}!({#4})$);
            \node[msgcircle, yshift=6.0mm] at (anchor) {#6};
            \node[yshift=2.0mm] at (anchor) {$\rightarrow$};
      }{}

      \ifthenelse{\isin{#1}{down} \AND \isin{#2}{left}}{
            \coordinate (anchor) at ($({#3})!{#5}!({#4})$);
            \node[msgcircle, yshift=-6.0mm] at (anchor) {#6};
            \node[yshift=-2.0mm] at (anchor) {$\leftarrow$};
      }{}
      \ifthenelse{\isin{#1}{up} \AND \isin{#2}{left}}{
            \coordinate (anchor) at ($({#3})!{#5}!({#4})$);
            \node[msgcircle, yshift=6.0mm] at (anchor) {#6};
            \node[yshift=2.0mm] at (anchor) {$\leftarrow$};
      }{}

      \ifthenelse{\isin{#1}{left} \AND \isin{#2}{up}}{
            \coordinate (anchor) at ($({#3})!{#5}!({#4})$);
            \node[msgcircle, xshift=-5.5mm] at (anchor) {#6};
            \node[xshift=-1.5mm] at (anchor) {$\uparrow$};
      }{}
      \ifthenelse{\isin{#1}{right} \AND \isin{#2}{up}}{
            \coordinate (anchor) at ($({#3})!{#5}!({#4})$);
            \node[msgcircle, xshift=5.5mm] at (anchor) {#6};
            \node[xshift=1.5mm] at (anchor) {$\uparrow$};
      }{}
}

\newcommand{\darkmsg}[6]{
      \ifthenelse{\isin{#1}{left} \AND \isin{#2}{down}}{
            \coordinate (anchor) at ($({#3})!{#5}!({#4})$);
            \node[darkmsgcircle, xshift=-5.5mm] at (anchor) {#6};
            \node[xshift=-1.5mm] at (anchor) {$\downarrow$};
      }{}
      \ifthenelse{\isin{#1}{right} \AND \isin{#2}{down}}{
            \coordinate (anchor) at ($({#3})!{#5}!({#4})$);
            \node[darkmsgcircle, xshift=5.5mm] at (anchor) {#6};
            \node[xshift=1.5mm] at (anchor) {$\downarrow$};
      }{}

      \ifthenelse{\isin{#1}{down} \AND \isin{#2}{right}}{
            \coordinate (anchor) at ($({#3})!{#5}!({#4})$);
            \node[darkmsgcircle, yshift=-6.0mm] at (anchor) {#6};
            \node[yshift=-2.0mm] at (anchor) {$\rightarrow$};
      }{}
      \ifthenelse{\isin{#1}{up} \AND \isin{#2}{right}}{
            \coordinate (anchor) at ($({#3})!{#5}!({#4})$);
            \node[darkmsgcircle, yshift=6.0mm] at (anchor) {#6};
            \node[yshift=2.0mm] at (anchor) {$\rightarrow$};
      }{}

      \ifthenelse{\isin{#1}{down} \AND \isin{#2}{left}}{
            \coordinate (anchor) at ($({#3})!{#5}!({#4})$);
            \node[darkmsgcircle, yshift=-6.0mm] at (anchor) {#6};
            \node[yshift=-2.0mm] at (anchor) {$\leftarrow$};
      }{}
      \ifthenelse{\isin{#1}{up} \AND \isin{#2}{left}}{
            \coordinate (anchor) at ($({#3})!{#5}!({#4})$);
            \node[darkmsgcircle, yshift=6.0mm] at (anchor) {#6};
            \node[yshift=2.0mm] at (anchor) {$\leftarrow$};
      }{}

      \ifthenelse{\isin{#1}{left} \AND \isin{#2}{up}}{
            \coordinate (anchor) at ($({#3})!{#5}!({#4})$);
            \node[darkmsgcircle, xshift=-5.5mm] at (anchor) {#6};
            \node[xshift=-1.5mm] at (anchor) {$\uparrow$};
      }{}
      \ifthenelse{\isin{#1}{right} \AND \isin{#2}{up}}{
            \coordinate (anchor) at ($({#3})!{#5}!({#4})$);
            \node[darkmsgcircle, xshift=5.5mm] at (anchor) {#6};
            \node[xshift=1.5mm] at (anchor) {$\uparrow$};
      }{}
}

\newcommand{\bwmsg}[6]{
      \ifthenelse{\isin{#1}{left} \AND \isin{#2}{down}}{
            \coordinate (anchor) at ($({#3})!{#5}!({#4})$);
            \node[msgdoublecircle, xshift=-5.5mm] at (anchor) {#6};
            \node[xshift=-1.5mm] at (anchor) {$\downarrow$};
      }{}
      \ifthenelse{\isin{#1}{right} \AND \isin{#2}{down}}{
            \coordinate (anchor) at ($({#3})!{#5}!({#4})$);
            \node[msgdoublecircle, xshift=5.5mm] at (anchor) {#6};
            \node[xshift=1.5mm] at (anchor) {$\downarrow$};
      }{}

      \ifthenelse{\isin{#1}{down} \AND \isin{#2}{right}}{
            \coordinate (anchor) at ($({#3})!{#5}!({#4})$);
            \node[msgdoublecircle, yshift=-6.0mm] at (anchor) {#6};
            \node[yshift=-2.0mm] at (anchor) {$\rightarrow$};
      }{}
      \ifthenelse{\isin{#1}{up} \AND \isin{#2}{right}}{
            \coordinate (anchor) at ($({#3})!{#5}!({#4})$);
            \node[msgdoublecircle, yshift=6.0mm] at (anchor) {#6};
            \node[yshift=2.0mm] at (anchor) {$\rightarrow$};
      }{}

      \ifthenelse{\isin{#1}{down} \AND \isin{#2}{left}}{
            \coordinate (anchor) at ($({#3})!{#5}!({#4})$);
            \node[msgdoublecircle, yshift=-6.0mm] at (anchor) {#6};
            \node[yshift=-2.0mm] at (anchor) {$\leftarrow$};
      }{}
      \ifthenelse{\isin{#1}{up} \AND \isin{#2}{left}}{
            \coordinate (anchor) at ($({#3})!{#5}!({#4})$);
            \node[msgdoublecircle, yshift=6.0mm] at (anchor) {#6};
            \node[yshift=2.0mm] at (anchor) {$\leftarrow$};
      }{}

      \ifthenelse{\isin{#1}{left} \AND \isin{#2}{up}}{
            \coordinate (anchor) at ($({#3})!{#5}!({#4})$);
            \node[msgdoublecircle, xshift=-5.5mm] at (anchor) {#6};
            \node[xshift=-1.5mm] at (anchor) {$\uparrow$};
      }{}
      \ifthenelse{\isin{#1}{right} \AND \isin{#2}{up}}{
            \coordinate (anchor) at ($({#3})!{#5}!({#4})$);
            \node[msgdoublecircle, xshift=5.5mm] at (anchor) {#6};
            \node[xshift=1.5mm] at (anchor) {$\uparrow$};
      }{}
}

\newcommand{\bwdarkmsg}[6]{
      \ifthenelse{\isin{#1}{left} \AND \isin{#2}{down}}{
            \coordinate (anchor) at ($({#3})!{#5}!({#4})$);
            \node[darkmsgdoublecircle, xshift=-5.5mm] at (anchor) {#6};
            \node[xshift=-1.5mm] at (anchor) {$\downarrow$};
      }{}
      \ifthenelse{\isin{#1}{right} \AND \isin{#2}{down}}{
            \coordinate (anchor) at ($({#3})!{#5}!({#4})$);
            \node[darkmsgdoublecircle, xshift=5.5mm] at (anchor) {#6};
            \node[xshift=1.5mm] at (anchor) {$\downarrow$};
      }{}

      \ifthenelse{\isin{#1}{down} \AND \isin{#2}{right}}{
            \coordinate (anchor) at ($({#3})!{#5}!({#4})$);
            \node[darkmsgdoublecircle, yshift=-6.0mm] at (anchor) {#6};
            \node[yshift=-2.0mm] at (anchor) {$\rightarrow$};
      }{}
      \ifthenelse{\isin{#1}{up} \AND \isin{#2}{right}}{
            \coordinate (anchor) at ($({#3})!{#5}!({#4})$);
            \node[darkmsgdoublecircle, yshift=6.0mm] at (anchor) {#6};
            \node[yshift=2.0mm] at (anchor) {$\rightarrow$};
      }{}

      \ifthenelse{\isin{#1}{down} \AND \isin{#2}{left}}{
            \coordinate (anchor) at ($({#3})!{#5}!({#4})$);
            \node[darkmsgdoublecircle, yshift=-6.0mm] at (anchor) {#6};
            \node[yshift=-2.0mm] at (anchor) {$\leftarrow$};
      }{}
      \ifthenelse{\isin{#1}{up} \AND \isin{#2}{left}}{
            \coordinate (anchor) at ($({#3})!{#5}!({#4})$);
            \node[darkmsgdoublecircle, yshift=6.0mm] at (anchor) {#6};
            \node[yshift=2.0mm] at (anchor) {$\leftarrow$};
      }{}

      \ifthenelse{\isin{#1}{left} \AND \isin{#2}{up}}{
            \coordinate (anchor) at ($({#3})!{#5}!({#4})$);
            \node[darkmsgdoublecircle, xshift=-5.5mm] at (anchor) {#6};
            \node[xshift=-1.5mm] at (anchor) {$\uparrow$};
      }{}
      \ifthenelse{\isin{#1}{right} \AND \isin{#2}{up}}{
            \coordinate (anchor) at ($({#3})!{#5}!({#4})$);
            \node[darkmsgdoublecircle, xshift=5.5mm] at (anchor) {#6};
            \node[xshift=1.5mm] at (anchor) {$\uparrow$};
      }{}
}
\usepackage[utf8]{inputenc}
\usepackage{array}
\usepackage{graphicx}
\usepackage{xcolor}
\usepackage{placeins}
\usepackage[hyphens]{url}
\usepackage{todonotes}
\usepackage{amsmath}
\usepackage{caption}
\usepackage{subcaption}
\usepackage[T1]{fontenc}
\usepackage[frozencache]{minted}
\usepackage{newverbs}

\definecolor{LightGray}{gray}{0.95}

\newverbcommand{\bverb}
    {\begin{lrbox}{\verbbox}}
    {\end{lrbox}\colorbox{LightGray}{\box\verbbox}}

\usepackage{amsfonts,amsmath,amssymb}
\renewcommand{\d}[1]{\operatorname{d}\!{#1}}
\renewcommand{\exp}[1]{\operatorname{exp}\!\left({#1}\right)}
\newcommand{\lognb}[1]{\operatorname{log}{#1}}

\newcommand{\T}{\operatorname{T}}

\newcommand{\vect}[1]{\boldsymbol{#1}}
\newcommand{\matr}[1]{\mathbf{#1}}

\newcommand{\N}[1]{\mathcal{N}\!\left({#1}\right)}

\usepackage[symbol]{footmisc}
\renewcommand{\thefootnote}{\fnsymbol{footnote}}

\journal{International Journal on Approximate Reasoning}

\begin{document}

\begin{frontmatter}
\title{A Factor Graph Approach to Automated Design of\\ Bayesian Signal Processing Algorithms}
\author[tue]{Marco Cox\footnote[1]{Joint first authors, order decided by coin toss.}$^{,}$}
\author[tue]{Thijs van de Laar\footnotemark[1]$^{,}$}
\author[tue,gn]{Bert de Vries}
\address[tue]{Department of Electrical Engineering, Eindhoven University of Technology, PO Box 513, 6500 MB, Eindhoven, the Netherlands}
\address[gn]{GN Hearing, Het Eeuwsel 6, 5612 AS, Eindhoven, The Netherlands}
\begin{keyword}
    Probabilistic Programming \sep Bayesian Inference \sep Julia \sep Factor Graphs \sep Message Passing
\end{keyword}

\begin{abstract}
The benefits of automating design cycles for Bayesian inference-based algorithms are becoming increasingly recognized by the machine learning community. As a result, interest in probabilistic programming frameworks has much increased over the past few years. This paper explores a specific probabilistic programming paradigm, namely message passing in Forney-style factor graphs (FFGs), in the context of automated design of efficient Bayesian signal processing algorithms. To this end, we developed ``ForneyLab''\footnote[2]{ForneyLab is available for download at \url{https://github.com/biaslab/ForneyLab.jl}.} as a Julia toolbox for message passing-based inference in FFGs. We show by example how ForneyLab enables automatic derivation of Bayesian signal processing algorithms, including algorithms for parameter estimation and model comparison. Crucially, due to the modular makeup of the FFG framework, both the model specification and inference methods are readily extensible in ForneyLab. In order to test this framework, we compared variational message passing as implemented by ForneyLab with automatic differentiation variational inference (ADVI) and Monte Carlo methods as implemented by state-of-the-art tools ``Edward'' and ``Stan''. In terms of performance, extensibility and stability issues, ForneyLab appears to enjoy an edge relative to its competitors for automated inference in state-space models.
\end{abstract}
\end{frontmatter}

\renewcommand{\thefootnote}{\fnsymbol{footnote}}

\section{Introduction}
The design of signal processing algorithms by probabilistic modeling comprises an iterative process that involves three phases: (1) model specification, (2) probabilistic inference (i.e., the actual algorithm derivation) and (3) performance evaluation (i.e., scoring of the algorithm). In this framework, a (signal processing) algorithm is defined as an inference task on a probabilistic model. For example, a Kalman filter-based algorithm can be specified as an inference task on a linear Gaussian dynamical system.

Based on the algorithm scoring results (phase 3), one might revise the model specification and repeat the process. In \cite{blei_build_2014}, this ``build, compute, critique, repeat''-cycle is called ``Box's loop'' and a strong argument can be made that this iterative process realizes the general scientific method. The great promise of a probabilistic modeling approach to algorithm design is that both the inference and scoring phases (phases 2 and 3) are results of Bayesian inference and therefore in principle automatable. If indeed phases 2 and 3 were automated by a suitable software suite, then a (human) algorithm designer could quickly loop through design iterations by proposing alternative models until a satisfactory performance score has been reached.

In practice, fully automating ``inference'' and ``scoring'' phases by a Bayesian inference toolbox is a yet unsolved problem. This has a limiting effect on the number of affordable iterations through Box's loop, which ultimately limits the quality of the final result.

In order to reduce the time and effort spent in the ``inference'' and ``scoring'' phases, an extensive line of research has focused on automating the derivation and implementation of (Bayesian) inference algorithms, dating back (at least) to the BUGS project that started in 1989 \cite{lunn_winbugs_2000}. The general idea behind this \textit{probabilistic programming} approach is to develop software that accepts a probabilistic model specification and returns an (approximate) Bayesian inference algorithm, without the need for manual derivations. Historically, probabilistic programming systems have relied heavily on Markov chain Monte Carlo (MCMC) methods due to their broad applicability. More recently, techniques like black-box variational inference (BBVI) \cite{ranganath_black_2014} have been added to the mix, for example in the popular probabilistic programming packages Stan \cite{carpenter_stan:_2017} and Edward \cite{tran_edward:_2016}.

Automatic generation of probabilistic inference algorithms involves an important trade-off between generality and efficiency. Inference methods that can automatically be applied to a wide array of models are usually not the most efficient, due to their black-box nature that prevents exploitation of model-specific properties. For example, MCMC methods are very generic, but can be orders of magnitude slower than inference algorithms that exploit model-specific properties such as conjugacy. This makes MCMC-based methods less suitable for situations that require real-time data processing or setups with limited computational resources.
On the other hand, computationally more efficient methods such as exact Bayesian inference, variational Bayesian inference and expectation propagation require model-specific derivations, which makes it harder to generate them automatically for arbitrary models. Recent work has focused on the design of Bayesian inference algorithms that are more efficient than vanilla MCMC methods while still being broadly applicable \cite{ranganath_black_2014,lienart_expectation_2015,jitkrittum_kernel-based_2015}.

In this paper we focus on message passing in factor graphs as a platform for automated design of Bayesian inference algorithms. Message passing exploits local model structure, while retaining general applicability. For instance, inference algorithms such as belief propagation \cite{pearl_reverend_1982}, variational Bayes \cite{winn_variational_2005}, expectation propagation \cite{minka_expectation_2001} and particle filtering \cite{dauwels_particle_2006} have already been formulated as message passing algorithms. The appeal of message passing is mainly due to its divide-and-conquer approach to inference, which allows it to marry the computational efficiency of analytic methods with the generality of black-box methods.

The goal of this paper is to paint a spectrum of possibilities that arise when adhering to the message passing approach to Bayesian inference, with a focus on time series modeling. Throughout the paper we present concrete examples that are implemented with ForneyLab, a novel publicly available toolbox we developed for generating message passing algorithms on Forney-style factor graphs \cite{forney_codes_2001}.

After a short technical introduction (Sec.~\ref{sec:mp_intro}), we illustrate how the message passing approach to inference with ForneyLab enables
\begin{itemize}
    \item automated design of message passing algorithms (Sec.~\ref{sec:automated});
    \item effective and flexible model design (Sec.~\ref{sec:effective_flexible});
    \item efficient Bayesian inference (Sec.~\ref{sec:efficient}).
\end{itemize}
Finally, we discuss related work (Sec.~\ref{sec:related_work}), and conclude (Sec.~\ref{sec:discussion}) by connecting ideas from the literature with the present framework.

\section{Background: Forney-style factor graphs and message passing algorithms}
\label{sec:mp_intro}

This section provides a short technical summary of message passing-based inference on Forney-style factor graphs. A more extensive introduction is available in \cite{loeliger_introduction_2004}. Furthermore, a detailed description of message passing on Forney-style factor graphs in the context of signal processing is available in \cite{korl_factor_2005}.

\subsection{Message passing on Forney-style factor graphs}
\begin{figure}[h]
    \center
    \begin{tikzpicture}
    [node distance=20mm,auto,>=stealth']


    \node[box] (f_b) {$f_b$};
    \node[box, left of=f_b] (f_a) {$f_a$};
    \node[box, right of=f_b] (f_c) {$f_c$};
    \node[right of=f_c, node distance=15mm] (f_c_r) {};

    \node[box, below of=f_c] (f_d) {$f_d$};
    \node[below of=f_d, node distance=15mm] (f_d_b) {};

    \path[line] (f_a) edge[->] node[anchor=south]{$x_1$} (f_b);
    \path[line] (f_b) edge[->] node[anchor=south]{$x_2$} (f_c);
    \path[line] (f_c) edge[->] node[anchor=south]{$x_3$} (f_c_r);

    \path[line] (f_c) edge[->] node[anchor=east]{$x_4$} (f_d);
    \path[line] (f_d) edge[->] node[anchor=east]{$x_5$} (f_d_b);
\end{tikzpicture}
    \caption{Forney-style factor graph (FFG) representation of Eq.~\ref{eq:example_gm}. In an FFG, edges correspond to variables and nodes represent factors that encode constraints among variables. A node connects to all edges that correspond to variables that occur in its factor function. For example, node $f_b$ connects to edges $x_1$ and $x_2$ since those variables occur in $f_b(x_1, x_2)$. Variables that occur in just one factor ($x_3$ and $x_5$ in this case) are represented by half-edges. While an FFG is principally an undirected graph, we usually specify a direction for the (half-)edges to indicate the generative direction of the model and to anchor the direction of messages flowing on the graph.}
    \label{fig:example_ffg}
\end{figure}
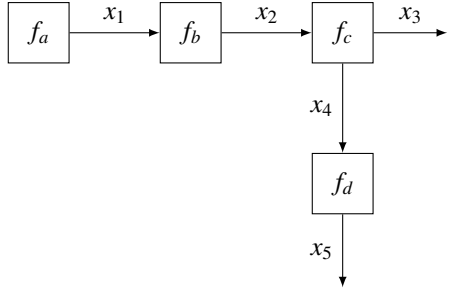

A Forney-style factor graph (FFG) \cite{forney_codes_2001} offers a graphical representation of a factorized probabilistic model. In an FFG, edges represent variables and nodes specify relations between variables. As a simple example, consider a generative model (joint probability distribution) over variables $x_1, \dots, x_5$ that factors as
\begin{align}
    f(x_1, \dots, x_5) = f_a(x_1) \, f_b(x_1, x_2)\, f_c(x_2, x_3, x_4)\, f_d(x_4, x_5)\,, \label{eq:example_gm}
\end{align}
where $f_\bullet(\cdot)$ denotes a probability density function. This factorized model can be represented graphically as an FFG, as shown in Fig.~\ref{fig:example_ffg}.
Note that although an FFG is principally an undirected graph, in the case of generative models we specify a direction for the edges to indicate the ``generative direction''. The edge direction simply anchors the direction of messages flowing on the graph (we speak of forward and backward messages that flow with or against the edge direction, respectively). In other words, the edge directionality is purely a notational issue and has no computational consequences.

The FFG representation of a probabilistic model helps to automate probabilistic inference tasks. As an example, consider we observe $x_5 = \hat{x}_5$ and are interested in calculating the marginal posterior probability distribution of $x_2$ given this observation.

\begin{figure}[ht]
    \center
    \begin{tikzpicture}
    [node distance=20mm,auto,>=stealth']


    \node[box] (f_b) {$f_b$};
    \node[box, left of=f_b] (f_a) {$f_a$};
    \node[box, right of=f_b, node distance=25mm] (f_c) {$f_c$};
    \node[right of=f_c, node distance=15mm] (f_c_r) {};

    \node[box, below of=f_c] (f_d) {$f_d$};
    \node[clamped, below of=f_d, node distance=15mm] (f_d_b) {};

    \path[line] (f_a) edge[->] node[anchor=north]{$x_1$} (f_b);
    \path[line] (f_b) edge[->] node[anchor=south, pos=0.2]{$x_2$} (f_c);
    \path[line] (f_c) edge[->] node[anchor=south]{$x_3$} node[anchor=north]{$\leftarrow$} (f_c_r);

    \path[line] (f_c) edge[->] node[anchor=east]{$x_4$} (f_d);
    \path[line] (f_d) edge[->] node[anchor=east]{$x_5$} node[anchor=west]{$\uparrow$} (f_d_b);

    \msg{up}{right}{f_a}{f_b}{0.5}{1};
    \msg{up}{right}{f_b}{f_c}{0.54}{2};
    \msg{right}{up}{f_d}{f_c}{0.5}{3};
    \msg{down}{left}{f_b}{f_c}{0.54}{4};

    \draw[dashed] (-2.6, 0.6) rectangle (-1.4, -0.6);
    \draw[dashed] (-2.8, 1.0) rectangle (1.0, -0.8);
    \draw[dashed] (1.9, -1.4) rectangle (3.1, -3.8);
    \draw[dashed] (1.7, 0.6) rectangle (4.1, -4.0);
\end{tikzpicture}
    \caption{Visualization of the message passing schedule corresponding to Eq.~\ref{eq:example_messages} with observed variable $x_5=\hat{x}_5$. The observation is indicated by terminating edge $x_5$ by a small solid node that technically represents the factor $\delta(x_5-\hat{x}_5)$. Messages are represented by numbered arrows, and the message sequence is chosen such that there are only backward dependencies. Dashed boxes mark the parts of the graph that are covered by the respective messages coming out of those boxes. The marginal posterior distribution $f(x_2 \mid x_5=\hat{x}_5)$ is obtained by taking the product of the messages that flow on edge $x_2$ and normalizing.}
    \label{fig:example_schedule}
\end{figure}
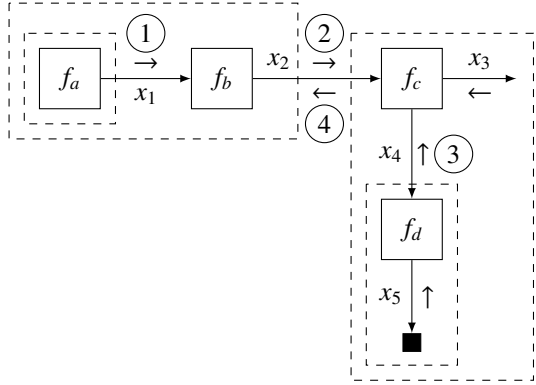

In the FFG context, observing the realization of a variable leads to the introduction of an extra factor in the model which ``clamps'' the variable to its observed value. In our example where $x_5$ is observed at value $\hat{x}_5$, we extend the generative model to $f(x_1,\ldots,x_5)\cdot \delta(x_5-\hat{x}_5)$. Following the notation introduced in \cite{reller_state-space_2012}, we denote such ``clamping'' factors in the FFG by solid black nodes. The FFG of the extended model is illustrated in Fig.~\ref{fig:example_schedule}.

Computing the marginal posterior distribution of $x_2$ under the observation $x_5 = \hat{x}_5$ involves integrating the extended model over all variables except $x_2$, and renormalizing:
\begin{subequations}
\begin{align}
    f(x_2 \mid x_5=\hat{x}_5) &\propto \idotsint f(x_1, \dots, x_5) \cdot \delta(x_5-\hat{x}_5) \d{x_1}\d{x_3}\d{x_4}\d{x_5} \label{eq:example_inference_def}\\
    &= \overbrace{\int \underbrace{f_a(x_1)}_{\circled{1}} \, f_b(x_1, x_2) \d{x_1}}^{\circled{2}} \, \overbrace{\iint f_c(x_2, x_3, x_4)\, \underbrace{\left(\int f_d(x_4, x_5) \cdot \delta(x_5-\hat{x}_5) \d{x_5}\right)}_{\circled{3}} \d{x_3}\d{x_4}}^{\circled{4}} \label{eq:example_messages}\,.
\end{align}
\end{subequations}
The nested integrals in Eq.~\ref{eq:example_messages} result from substituting the factorization of Eq.~\ref{eq:example_gm} and rearranging the integrals according to the distributive law. Rearranging large integrals of this type as a product of nested sub-integrals can be automated by exploiting the FFG representation of the corresponding model. The sub-integrals indicated by circled numbers correspond to integrals over parts of the model (indicated by dashed boxes in Fig.~\ref{fig:example_schedule}), and their solutions can be interpreted as messages flowing on the FFG. Therefore, this procedure is known as \textit{message passing} (or summary propagation). The messages are ordered (``scheduled'') in such a way that there are only backward dependencies, i.e., each message can be calculated from preceding messages in the schedule. Crucially, these schedules can be generated automatically, for example by performing a depth-first search on the FFG.

Message passing is generally efficient because the computation of every message is node-local in the FFG. More specifically, the message flowing out of a factor node $f_a$ can be calculated from the analytic form of factor $f_a$ and all messages inbound to node $f_a$. If the analytic forms of the incoming messages are known (which is often the case), a pre-derived \textit{message computation rule} can be used to compute the outgoing message. These rules can be stored in a lookup table for reuse in any model that involves that specific factor-message combination. This important locality property thus enables efficient and automated probabilistic inference.

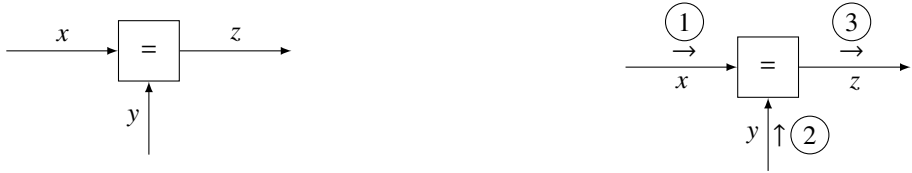
\begin{figure}[ht]
    \begin{subfigure}{.49\textwidth}
        \center
        \begin{tikzpicture}
    [node distance=20mm,auto,>=stealth']


    \node[box] (eq) {$=$};

    \node[left of=eq] (x) {};
    \node[below of=eq, node distance=15mm] (y) {};
    \node[right of=eq] (z) {};

    \path[line] (x) edge[->] node[anchor=south]{$x$} (eq);
    \path[line] (y) edge[->] node[anchor=east]{$y$} (eq);
    \path[line] (eq) edge[->] node[anchor=south]{$z$} (z);
\end{tikzpicture}
    \end{subfigure}
    \begin{subfigure}{.49\textwidth}
        \center
        \begin{tikzpicture}
    [node distance=20mm,auto,>=stealth']


    \node[box] (eq) {$=$};

    \node[left of=eq] (x) {};
    \node[below of=eq, node distance=15mm] (y) {};
    \node[right of=eq] (z) {};

    \path[line] (x) edge[->] node[anchor=north]{$x$} (eq);
    \path[line] (y) edge[->] node[anchor=east]{$y$} (eq);
    \path[line] (eq) edge[->] node[anchor=north]{$z$} (z);

    \msg{up}{right}{x}{eq}{0.45}{1};
    \msg{right}{up}{y}{eq}{0.4}{2};
    \msg{up}{right}{eq}{z}{0.55}{3};
\end{tikzpicture}
    \end{subfigure}
    \caption{FFG representation (left) and message passing schedule (right) for an equality constraint node.}
    \label{fig:equality}
\end{figure}

In the case of marginalization, the messages are derived according to the so-called \textit{sum-product rule}\footnote[3]{The name sum-product rule derives from the observation that each sub-integral in Eq.~\ref{eq:example_messages} comprises a sum (integral) of a product of factors.}, which leads to the sum-product (belief propagation) algorithm. As an example derivation we consider the outgoing message of an ``equality constraint node'' (see Fig.~\ref{fig:equality}, left; see also \cite{korl_factor_2005}), which constrains three variables $x, y, z$ to equal values through the factor $f_{=}(x, y, z) = \delta(z - x)\, \delta(z - y)$.

For given incoming messages $\mu_1(x)$ and $\mu_2(y)$ on edges $x$ and $y$ (depicted by $\circled{1}$ and $\circled{2}$ in the Fig.~\ref{fig:equality}, right), the outgoing sum-product message on the $z$-edge is given by
\begin{subequations}
\begin{align}
    \mu_3(z) &= \iint \mu_1(x) \, \mu_2(y)\, f_{=}(x, y, z) \d{x}\d{y} \label{eq:equality-a}\\
    &= \mu_1(z) \, \mu_2(z)\,. \label{eq:equality-b}
\end{align}
\end{subequations}
Note that the outgoing message is only a function of $z$ and that it is calculated from only node-local information, namely the incoming messages at the corresponding node and the definition of the node factor itself. Equality constraint nodes are quite prevalent in FFGs because they constitute a branching mechanism that distributes variables over multiple (more than two) factors in the graph. If we interpret message $\mu_1(\cdot)$ as a prior and message $\mu_2(\cdot)$ as a likelihood function, then message $\mu_3(\cdot)$ becomes proportional to the posterior distribution over $z$. Therefore, message passing through the equality node effectively fuses information from two sources by executing Bayes rule (up to a normalizing constant).

\subsection{Variational message passing}\label{sec:variational-message-passing}
Inference problems on practical models often lead to sub-integrals that are difficult to evaluate analytically. In such cases, exact inference through sum-product message passing is impractical. However, one can often resort to alternative message passing algorithms that yield approximate solutions. One such algorithm is variational message passing (VMP) \cite{winn_variational_2005, dauwels_variational_2007}, which will be applied extensively throughout this paper.

VMP is the message passing implementation of variational Bayesian inference, which finds an approximate solution to an inference problem by reformulating inference as an optimization task \cite{attias_variational_1999}. Concretely, for a given model $p(\vect{y},\vect{z})$ with hidden variables $\vect{z}$ and observed variables $\vect{y}$, we define an approximate inference solution $q(\vect{z}) \approx p(\vect{z}\,|\,\vect{y})$ (also known as the \emph{recognition} distribution), and a so-called variational free energy functional
\begin{subequations}
\begin{align}
F[q] &\triangleq \underbrace{-\int_{\vect{z}} q(\vect{z}) \log p(\vect{y},\vect{z})\,\mathrm{d}\vect{z}}_{\text{energy}}  + \underbrace{\int_{\vect{z}} q(\vect{z}) \log q(\vect{z})\,\mathrm{d}\vect{z}}_{-\text{entropy}} \label{eq:FE-energy-entropy} \\
  &= \underbrace{-\log p(\vect{y})}_{-\text{log-evidence}} + \underbrace{\int_{\vect{z}} q(\vect{z}) \log \frac{q(\vect{z})}{p(\vect{z}|\vect{y})}\,\mathrm{d}\vect{z}}_{\text{KL-divergence}}\,. \label{eq:FE-evidence-divergence}
\end{align}
\end{subequations}
In the machine learning community, the \emph{negative} free energy ($-F[q]$) is also known as the Evidence Lower BOund (ELBO) \cite{blei_variational_2017}.
The variational Bayes algorithm proceeds by minimizing $F[q]$ with respect to the parameters of the approximate posterior $q(\vect{z})$ through some optimization procedure. Since only the second term in Eq.~\ref{eq:FE-evidence-divergence} involves $q(\vect{z})$, this is equivalent to minimizing the Kullback-Leibler (KL) divergence between the proposed solution $q(\vect{z})$ and the (perfect) Bayesian solution $p(\vect{z}|\vect{y})$. As a result, the approximate solution $q(\vect{z})$ is optimized to be as close as possible to the exact solution $p(\vect{z}|\vect{y})$ in terms of KL-divergence. Note also that if the minimized KL-divergence is small in comparison to the first term (minus-log-evidence), then the optimized free energy is a good approximation of the Bayesian model evidence. In practice, $q^*=\arg\min F[q]$ is used as a proxy for the target posterior, and $F[q^*]$ is often used to score the model performance. The analytic form of the approximate posterior $q(\vect{z})$ is usually chosen such that it can provide a reasonable approximation to the true posterior while keeping the optimization problem tractable.

In \cite{dauwels_variational_2007}, it is shown that variational free energy minimization as outlined above can be implemented by message passing on the FFG representation of the generative model. Similar to sum-product message passing, VMP involves the evaluation of a sequence of messages, where each message is calculated from node-local information. Every VMP message update corresponds to a coordinate descent step on the variational free energy, and therefore multiple passes through the schedule converge the free energy to a local minimum.


\section{Toolbox-based automated design of message passing algorithms}
\label{sec:automated}

Message passing provides a convenient paradigm for automating the design of (Bayesian) inference algorithms. A variety of (approximate) inference algorithms can be formulated in terms of message passing, including exact Bayesian inference (sum-product message passing, belief propagation), variational Bayes (VMP), expectation maximization (a special case of VMP), expectation propagation and Gibbs sampling. In principle, these inference algorithms can be generated automatically by finding appropriate message passing schedules and using a library of pre-derived message update equations.
This section introduces ForneyLab, a newly developed open source toolbox for automatic generation of inference algorithm based on this paradigm. After a description of the toolbox, we demonstrate its core functionalities through an example application.

\subsection{ForneyLab: A toolbox for automating message passing-based inference}
To facilitate the automatic generation of message passing solutions to inference problems, we developed ForneyLab (\url{https://github.com/biaslab/ForneyLab.jl}), which at the moment of writing this paper is released at version 0.8. ForneyLab is written in the open source scientific programming language Julia, which enjoys a MATLAB-like syntax and native speed similar to compiled \emph{C} code \cite{bezanson_julia:_2017}. ForneyLab accepts a specification of the probabilistic model and inference task as inputs, and produces interpretable source code for the desired inference algorithm, thus enabling users to inspect, modify and debug the (message passing-based inference) implementation. This is achieved by scheduling messages on an FFG representation of the model at hand, and using a library of built-in update rules.


Constructing a message passing algorithm with ForneyLab consists of two main steps: specifying the probabilistic model, and defining an inference problem on this model. For the model definition step, ForneyLab provides a convenient domain-specific syntax that resembles notational conventions of alternative probabilistic programming languages. Under the hood, ForneyLab builds the corresponding FFG. Optionally, custom factor nodes and message update rules can be defined outside the framework and re-used in model construction. A node function may internally be represented by a factor graph itself, giving rise to so-called \emph{composite} nodes. This type of support for structural abstraction may be convenient for hierarchical model construction and is also beneficial for algorithmic efficiency. Composite nodes are discussed in more detail in Sec.~\ref{sec:composability}.

Once the model has been defined, the user specifies an inference task, which may correspond to (for instance) a signal processing task (by online state estimation) or a parameter estimation task. In general, inference involves finding the marginal posterior distributions of a subset of the model variables. ForneyLab uses the FFG representation of the probabilistic model to automatically derive a suitable message passing schedule, i.e., a sequence of message updates that realizes the requested inference task, and generates source code that implements this message passing algorithm. More specifically, the inference algorithm generation pipeline involves the following stages:
\begin{enumerate}
\item \textbf{Scheduling}. This step corresponds to finding a sequence of message updates that yields all required marginal distributions.
\item \textbf{Update rule selection and message type inference}.\\
At each factor node, multiple message update rules may be available. ForneyLab chooses the most appropriate rule based on inbound message types and outbound message requirements. ForneyLab contains a library of pre-computed update rules for built-in nodes, but can also use custom, user-defined update rules (see Sec.\ref{sec:composability} for an example).
\item \textbf{Code generation}.\\
The sequence of message updates is compiled to source code. Currently, ForneyLab comes with a Julia code generator, but additional code generation engines can be added, for example to generate \emph{C} code or to target computational frameworks like TensorFlow.
\end{enumerate}

Splitting the algorithm generation process into separate stages allows the user to inspect and modify intermediate constructs. For example, it is possible to use a handcrafted schedule instead of an automatically generated one, or to manually change which update rule is applied for a certain message.
Since the final result is inference source code, the user is free to probe or manually modify the inference algorithm at any level.

In summary, ForneyLab consists of the following four components: (i) a convenient domain-specific syntax for specifying probabilistic models; (ii) a library of commonly used factor nodes and corresponding message update implementations; (iii) automatic algorithm generators for belief propagation, variational message passing and expectation propagation; and (iv) an ``algorithm-to-code'' compiler which generates readable and debuggable inference source code.
To demonstrate how the algorithm generation process works in practice, we proceed by working out an example.

\subsection{Example: automated inference in a hidden Markov model with Gaussian mixture emissions}\label{sec:hybrid}
\begin{figure}[h]
    \center
    \begin{tikzpicture}
    [node distance=20mm,auto,>=stealth']


    \node[box] (T) {};
    \node[below of=T, node distance=8mm] (p_T) {$p(\vect{x}_t \,|\, \vect{x}_{t-1}, \theta)$};
    \node[left of=T] (x_t_min_dots) {$\dots$};
    \node[above of=T, node distance=15mm] (theta) {};
    \node[box, right of=T] (A_eq) {$=$};
    \node[right of=A_eq] (x_t_dots) {$\dots$};

    \node[box, below of=A_eq] (O) {};
    \node[right of=O, node distance=15mm] (p_O) {$p(\vect{y}_t \,|\, \vect{x}_t, \phi)$};
    \node[left of=O, node distance=15mm] (phi) {};
    \node[below of=O, node distance=15mm] (y_t) {};

    \path[line] (x_t_min_dots) edge[->] node[anchor=south]{$\vect{x}_{t-1}$} (T);
    \path[dashed] (theta) edge[->] node[anchor=east]{$\theta$} (T);
    \path[line] (T) edge[->] (A_eq);
    \path[line] (A_eq) edge[->] node[anchor=south]{$\vect{x}_t$} (x_t_dots);

    \path[line] (A_eq) edge[->] (O);
    \path[dashed] (phi) edge[->] node[anchor=south]{$\phi$} (O);
    \path[line] (O) edge[->] node[anchor=east]{$\vect{y}_t$} (y_t);
\end{tikzpicture}
    \caption{Factor graph representation of the state-space model of Eq.~\ref{eq:ssm_gm} (priors not drawn). The dots on the left and right sides indicate a repetition of the model section over time. Dashed edges indicate time-independent parameters that are equality constrained over time.}
    \label{fig:ssm_ffg}
\end{figure}
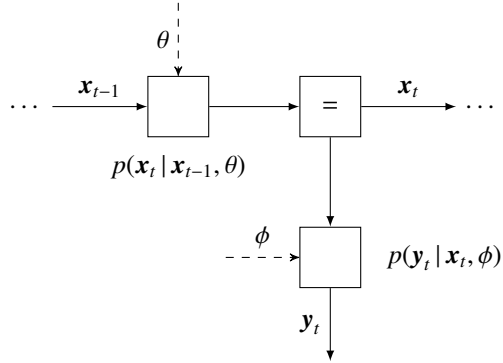

Message passing on factor graphs is known to support a very wide array of statistical signal processing, communication and control engineering algorithms, including Kalman filtering/smoothing, hidden Markov model learning, Viterbi decoding etc. \cite{loeliger_factor_2007}. Most of these algorithms can be interpreted as inference tasks on probabilistic \emph{state-space models} (SSM). In the following, we focus the discussion on automated toolbox-based derivation of inference tasks in probabilistic state-space models. A state-space model (SSM) is defined as
\begin{align}
    p(\vect{y}, \vect{x}, \theta, \phi) = \underbrace{p(\vect{x}_0) \, p(\theta) \, p(\phi)}_{\text{priors}} \prod_{t=1}^T \underbrace{p(\vect{x}_t \,|\, \vect{x}_{t-1}, \theta)}_{\text{state transition}} \, \underbrace{p(\vect{y}_t \,|\, \vect{x}_t, \phi)}_{\text{observation}}\,, \label{eq:ssm_gm}
\end{align}
where $\vect{x} = (\vect{x}_0, \vect{x}_1, \dots, \vect{x}_T)$ are hidden (unobserved) states, $\vect{y} = (\vect{y}_1, \dots, \vect{y}_T)$ are observed variables and $\{\theta, \phi\}$ collect the model parameters. In an FFG, time-independent model parameters are constrained to be equal over time by a chain of equality constraint nodes (one for each model section). In order to avoid cluttering the graph with equality chains, we denote these time-independent model parameters by dashed edges in the FFG. The general SSM is graphically represented by the FFG of Fig.~\ref{fig:ssm_ffg}.

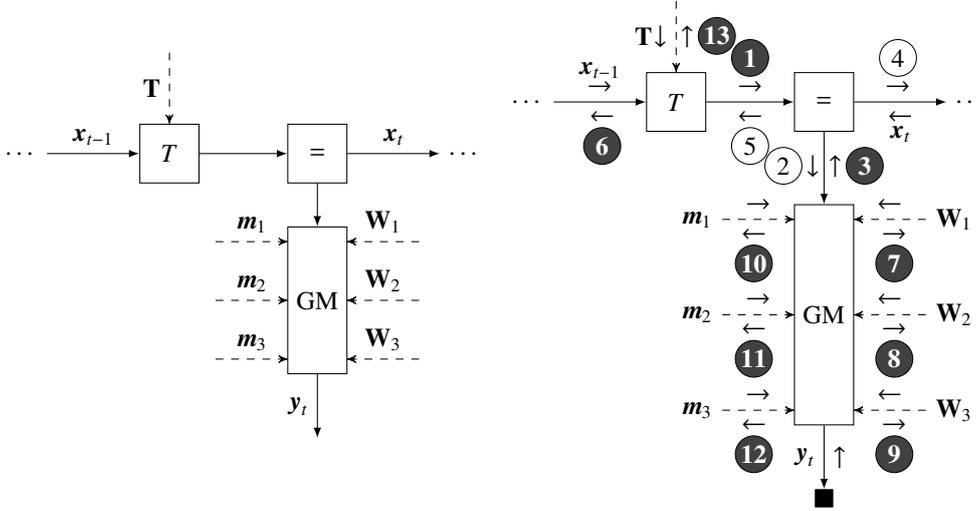
\begin{figure}[h]
    \center
    \begin{subfigure}{.4\textwidth}
        \center
        \resizebox{\textwidth}{!}{\begin{tikzpicture}
    [node distance=20mm,auto,>=stealth']


    \node[box] (AT) {$T$};
    \node[left of=AT] (x_t_min_dots) {$\dots$};
    \node[above of=AT, node distance=15mm] (A) {};
    \node[box, right of=AT] (A_eq) {$=$};
    \node[right of=A_eq] (x_t_dots) {$\dots$};

    \node[box, below of=A_eq, minimum height=2cm] (GM) {GM};
    \coordinate (GM_top_left) at ($(GM)+(-0.4,0.8)$);
    \node[left of=GM_top_left, node distance=11mm] (m_top_left) {};
    \coordinate (GM_middle_left) at ($(GM)+(-0.4,0.0)$);
    \node[left of=GM_middle_left, node distance=11mm] (m_middle_left) {};
    \coordinate (GM_bottom_left) at ($(GM)+(-0.4,-0.8)$);
    \node[left of=GM_bottom_left, node distance=11mm] (m_bottom_left) {};
    \coordinate (GM_top_right) at ($(GM)+(0.4,0.8)$);
    \node[right of=GM_top_right, node distance=11mm] (W_top_right) {};
    \coordinate (GM_middle_right) at ($(GM)+(0.4,0.0)$);
    \node[right of=GM_middle_right, node distance=11mm] (W_middle_right) {};
    \coordinate (GM_bottom_right) at ($(GM)+(0.4,-0.8)$);
    \node[right of=GM_bottom_right, node distance=11mm] (W_bottom_right) {};
    \node[below of=GM, node distance=20mm] (y_t) {};

    \path[line] (x_t_min_dots) edge[->] node[anchor=south]{$\vect{x}_{t-1}$} (AT);
    \path[dashed] (A) edge[->] node[anchor=east]{$\matr{T}$} (AT);
    \path[line] (AT) edge[->] (A_eq);
    \path[line] (A_eq) edge[->] node[anchor=south]{$\vect{x}_t$} (x_t_dots);

    \path[line] (A_eq) edge[->] (GM);
    \path[dashed] (m_top_left) edge[->] node[anchor=south]{$\vect{m}_1$} (GM_top_left);
    \path[dashed] (m_middle_left) edge[->] node[anchor=south]{$\vect{m}_2$} (GM_middle_left);
    \path[dashed] (m_bottom_left) edge[->] node[anchor=south]{$\vect{m}_3$} (GM_bottom_left);
    \path[dashed] (W_top_right) edge[->] node[anchor=south]{$\matr{W}_1$} (GM_top_right);
    \path[dashed] (W_middle_right) edge[->] node[anchor=south]{$\matr{W}_2$} (GM_middle_right);
    \path[dashed] (W_bottom_right) edge[->] node[anchor=south]{$\matr{W}_3$} (GM_bottom_right);
    \path[line] (GM) edge[->] node[anchor=east]{$\vect{y}_t$} (y_t);
\end{tikzpicture}}
    \end{subfigure}
    \begin{subfigure}{.4\textwidth}
        \center
        \resizebox{\textwidth}{!}{\begin{tikzpicture}
    [node distance=20mm,auto,>=stealth']


    \node[box] (AT) {$T$};
    \node[left of=AT] (x_t_min_dots) {$\dots$};
    \node[above of=AT, node distance=15mm] (A) {};
    \node[box, right of=AT] (A_eq) {$=$};
    \node[right of=A_eq] (x_t_dots) {$\dots$};

    \node[box, below of=A_eq, minimum height=3cm, node distance=29mm] (GM) {GM};
    \coordinate (GM_top_left) at ($(GM)+(-0.4,1.3)$);
    \node[left of=GM_top_left, node distance=11mm] (m_top_left) {};
    \coordinate (GM_middle_left) at ($(GM)+(-0.4,0.0)$);
    \node[left of=GM_middle_left, node distance=11mm] (m_middle_left) {};
    \coordinate (GM_bottom_left) at ($(GM)+(-0.4,-1.3)$);
    \node[left of=GM_bottom_left, node distance=11mm] (m_bottom_left) {};
    \coordinate (GM_top_right) at ($(GM)+(0.4,1.3)$);
    \node[right of=GM_top_right, node distance=11mm] (W_top_right) {};
    \coordinate (GM_middle_right) at ($(GM)+(0.4,0.0)$);
    \node[right of=GM_middle_right, node distance=11mm] (W_middle_right) {};
    \coordinate (GM_bottom_right) at ($(GM)+(0.4,-1.3)$);
    \node[right of=GM_bottom_right, node distance=11mm] (W_bottom_right) {};
    \node[clamped, below of=GM, node distance=25mm] (y_t) {};

    \path[line] (x_t_min_dots) edge[->] node[anchor=south, yshift=2mm]{$\vect{x}_{t-1}$} node[anchor=south]{$\rightarrow$} (AT);
    \path[dashed] (A) edge[->] node[anchor=east, xshift=-2mm]{$\matr{T}$} node[anchor=east]{$\downarrow$} (AT);
    \path[line] (AT) edge[->] (A_eq);
    \path[line] (A_eq) edge[->] node[anchor=north, yshift=-2mm]{$\vect{x}_t$} node[anchor=north]{$\leftarrow$} (x_t_dots);

    \path[line] (A_eq) edge[->] (GM);
    \path[dashed] (m_top_left) edge[->] node[anchor=east, xshift=-5mm]{$\vect{m}_1$} node[anchor=south]{$\rightarrow$} (GM_top_left);
    \path[dashed] (m_middle_left) edge[->] node[anchor=east, xshift=-5mm]{$\vect{m}_2$} node[anchor=south]{$\rightarrow$} (GM_middle_left);
    \path[dashed] (m_bottom_left) edge[->] node[anchor=east, xshift=-5mm]{$\vect{m}_3$} node[anchor=south]{$\rightarrow$} (GM_bottom_left);
    \path[dashed] (W_top_right) edge[->] node[anchor=west, xshift=5mm]{$\matr{W}_1$} node[anchor=south]{$\leftarrow$} (GM_top_right);
    \path[dashed] (W_middle_right) edge[->] node[anchor=west, xshift=5mm]{$\matr{W}_2$} node[anchor=south]{$\leftarrow$} (GM_middle_right);
    \path[dashed] (W_bottom_right) edge[->] node[anchor=west, xshift=5mm]{$\matr{W}_3$} node[anchor=south]{$\leftarrow$} (GM_bottom_right);
    \path[line] (GM) edge[->] node[anchor=east]{$\vect{y}_t$} node[anchor=west]{$\uparrow$} (y_t);

    \darkmsg{up}{right}{AT}{A_eq}{0.5}{1};
    \msg{left}{down}{A_eq}{GM}{0.3}{2};
    \darkmsg{right}{up}{GM}{A_eq}{0.7}{3};
    \msg{up}{right}{A_eq}{x_t_dots}{0.5}{4};
    \msg{down}{left}{A_eq}{AT}{0.5}{5};
    \darkmsg{down}{left}{AT}{x_t_min_dots}{0.5}{6};
    \darkmsg{down}{right}{GM_top_right}{W_top_right}{0.5}{7};
    \darkmsg{down}{right}{GM_middle_right}{W_middle_right}{0.5}{8};
    \darkmsg{down}{right}{GM_bottom_right}{W_bottom_right}{0.5}{9};
    \darkmsg{down}{left}{GM_top_left}{m_top_left}{0.5}{10};
    \darkmsg{down}{left}{GM_middle_left}{m_middle_left}{0.5}{11};
    \darkmsg{down}{left}{GM_bottom_left}{m_bottom_left}{0.5}{12};
    \darkmsg{right}{up}{AT}{A}{0.6}{13};
\end{tikzpicture}}
    \end{subfigure}
    \caption{Factor graph representation and VMP schedule for the HMGM model with $J=3$ components. Black-labeled messages are computed by the VMP update rule from \cite{dauwels_variational_2007} and white-labeled messages are calculated according to sum-product update rules.}
    \label{fig:dsco}
\end{figure}

As an example, we consider an SSM that combines a first-order Markov transition model for a discrete state (Eq.~\ref{eq:hybrid_transition}) with a continuously-valued Gaussian mixture observation model (Eq.~\ref{eq:hybrid_observation}). These types of models are successfully used in many applications such as the modeling of fabrication processes, behavioral data and speech signals. Specifically, we consider the model specified by
\begin{subequations}
\begin{align}
    p(\vect{x}_t \,|\, \vect{x}_{t-1}, \matr{T}) &= \textrm{Categorical}\left( \vect{x}_t \,\middle|\, \matr{T}\,\vect{x}_{t-1} \right) \label{eq:hybrid_transition},\\
    p(\vect{y}_t \,|\,\vect{x}_t, \vect{m}, \matr{W}) &= \prod_{k=1}^K \mathcal{N} \left( \vect{y}_t \,\middle|\, \vect{m}_k, \matr{W}_k^{-1} \right) ^{x_{t, k}}  \label{eq:hybrid_observation}\,,
\end{align}
\label{eq:hybrid}
\end{subequations}
where $\vect{x}_t$ is a one-hot coded vector representing the hidden state at time $t$, and $\matr{T}$ is the state transition probability matrix. The vector $\vect{y}_t \in \mathbb{R}^d$ holds the observations, and $\left\{\vect{m}_k, \matr{W}_k\right\}$ are the parameters for the $k$-th mixture component. 
The FFG for the generative model with $3$ components is drawn in Fig.~\ref{fig:dsco} (left). In contrast to a standard Gaussian mixture model (GMM), this model includes a time-dependent discrete state vector that identifies the mixture component from which the observations are drawn. Furthermore, the model differs from a standard hidden Markov model (HMM) because it includes a more complex emission model for continuously-valued observations (rather than discrete observations). We refer to this model as a Hidden Markov Gaussian Mixture (HMGM) model. In ForneyLab, the HMGM model is specified by the code fragment shown in Fig.~\ref{lst:build}.

\begin{figure}[ht]
\begin{minted}[baselinestretch=1.2,bgcolor=LightGray,fontsize=\footnotesize]{julia}
# Priors
@RV T ~ Dirichlet(ones(3,3))
@RV m1 ~ GaussianMeanVariance(zeros(2), huge*diageye(2))
@RV W1 ~ Wishart(huge*diageye(2), 2.0)
...
@RV x_0 ~ Categorical(1/3*ones(3), id=:x_0)

x = Vector{Variable}(n_samples) # Pre-allocate variable vector
y = Vector{Variable}(n_samples)
x_t_min = x_0 # Initialize previous state
for t = 1:n_samples # Build model sections
    @RV x[t] ~ Transition(x_t_min, T) # Transition model
    @RV y[t] ~ GaussianMixture(x[t], m1, W1, m2, W2, m3, W3) # Observation model

    x_t_min = x[t] # Reset state for next section

    placeholder(y[t], :y, index=t, datatype=Float64, dims=(2,)) # Indicate observation of y at time t
end
\end{minted}
\caption{Julia code for building the HMGM model from Eq.~\ref{eq:hybrid} with ForneyLab. In Julia, expressions prefixed by ``@'' indicate macros. Here, \texttt{@RV} describes a ``Random Variable''-node constructor. The ``:'' prefix (e.g. \texttt{:y}) identifies a symbol that may be used for indexing. The \texttt{placeholder} function marks a model variable as observed.}
\label{lst:build}
\end{figure}

Given a sequence of observations $\vect{y} = \left(\vect{y_1},\ldots,\vect{y_T}\right)$, we are interested in estimating both the hidden state sequence $\vect{x}$ and the model parameters $\{\matr{T},\vect{m},\matr{W}\}$. In other words, we wish to compute
\begin{align*}
    p(\vect{x}, \matr{T}, \vect{m}, \matr{W} \,|\, \vect{y}) &= \frac{p(\vect{y}, \vect{x}, \matr{T}, \vect{m}, \matr{W})}{\sum_{\vect{x}} \iiint p(\vect{y}, \vect{x}, \matr{T}, \vect{m}, \matr{W}) \d{\matr{T}} \d{\vect{m}} \d{\matr{W}}}\,.
\end{align*}
The integrals in the denominator are not analytically tractable, making it impossible to perform exact inference, for example through sum-product message passing. However, it is possible to find an approximate solution to the inference problem by resorting to the variational Bayes algorithm, which we will do here.
For the variational approximation to the true posterior, we choose the following factorization:
\begin{align}\label{eq:q-factorization}
    q(\vect{x}, \matr{T}, \vect{m}, \matr{W}) = q(\vect{x}) \, q(\matr{T}) \prod_{k=1}^K q(\vect{m}_k) \, q(\matr{W}_k)\,.
\end{align}
This factorization is known as a \textit{structured variational approximation} since $q(\vect{x})$ -- the approximate posterior distribution of the hidden state sequence -- does not fully factorize over all time-indexed state variables, i.e., we do not assume $q(\vect{x})=q(\vect{x_0}) q(\vect{x_1}) \cdots q(\vect{x_T})$. Given the factorization of Eq.~\ref{eq:q-factorization}, the optimal analytic forms of the factors of $q$ are determined by the generative model, so there is no need to specify these forms manually.

Fig.~\ref{lst:schedule} lists the code to specify the recognition distribution and build the corresponding inference algorithm with ForneyLab. The resulting message passing schedule is illustrated in Fig.~\ref{fig:dsco} (right) and involves both VMP (black-labeled) and sum-product (white-labeled) update rules. Where sum-product messages perform exact (Bayesian) updates, the VMP messages introduce approximations by directly using the recognition distributions in their update computations. For details, we refer to the detailed description of VMP on FFGs by Dauwels \cite{dauwels_variational_2007}.

The message passing schedule should be repeated until the free energy has converged to a local minimum. The source code that ForneyLab generates to implement the inference algorithm contains one (Julia) function for each factor in the recognition distribution. These functions update the parameters of the corresponding factor in the recognition distribution.
Fig.~\ref{lst:algo} shows a snippet of the function for updating $q(\vect{x})$. The complete message passing algorithm in this function contains 249 calls to message update functions. The snippet shows how the computation for \bverb|message[1]| depends on observation \bverb|data[:y][8]| and the current beliefs over the mixture components, as stored in the \bverb|marginals| dictionary. Furthermore, the computation for e.g. \bverb|message[10]| depends on \bverb|message[9]|, showing the sequentiality of the message passing algorithm. In the end, the resulting marginals, e.g. \bverb|marginals[:x_0]|, are updated by multiplying colliding messages. These marginals are then used in the updates for the other recognition factors.
Inspection of the (also automatically generated) algorithm for evaluating the variational free energy (Fig.~\ref{lst:algo_F}) reveals that the free energy is also computed by a sequence of node-local computations. As discussed in Sec.~\ref{sec:variational-message-passing}, the value of the minimized free energy functional can be used as a performance metric for the generated algorithm, since minimized free energy is a proxy for ``minus log-evidence''.

\begin{figure}[ht]
\begin{minted}[baselinestretch=1.2,bgcolor=LightGray,fontsize=\footnotesize]{julia}
q = RecognitionFactorization([x_0; x], T, m1, W1, m2, W2, m3, W3; ids=[:X, :T, :M1, :W1, :M2, :W2, :M3, :W3])
algo = variationalAlgorithm(q) # Build the VMP algorithm
algo_F = freeEnergyAlgorithm(q) # Build algorithm to evaluate the variational free energy
\end{minted}
\caption{Julia code for specifying and building a variational message passing algorithm with ForneyLab. The first line specifies a structured recognition distribution, where all state variables are accommodated in a single recognition factor (indicated by square brackets), with corresponding id \texttt{:X}. Subsequent variables each have their own recognition factor and corresponding ids. The final two lines invoke the ForneyLab algorithm generators that return the variational message passing algorithm and free energy algorithm as Julia source code.}
\label{lst:schedule}
\end{figure}
\begin{figure}[ht]
\begin{minted}[baselinestretch=1.2,bgcolor=LightGray,fontsize=\footnotesize]{julia}
function stepX!(data::Dict, marginals::Dict=Dict(), messages::Vector{Message}=Array{Message}(249))
    messages[1] = ruleVBGaussianMixtureZCat(ProbabilityDistribution(Multivariate, PointMass, m=data[:y][8]),
        nothing, marginals[:m1], marginals[:W1], marginals[:m2], marginals[:W2], marginals[:m3],
        marginals[:W3])
    ...
    messages[10] = ruleSVBTransitionOutVCD(nothing, messages[9], marginals[:T])
    ...
    marginals[:x_0] = messages[9].dist * messages[249].dist
    marginals[:x_1] = messages[10].dist * messages[248].dist
    ...
    return marginals
end
\end{minted}
\caption{Segment of an automatically generated inference algorithm code as stored in \texttt{algo} (Fig.~\ref{lst:schedule}). For variational message passing, each recognition factor has its own corresponding \texttt{step!} function, where the trailing characters relate to the corresponding recognition factor, e.g. \texttt{stepX!}. Each \texttt{step!} function consecutively builds an array of messages by executing specific message update rules. In this example, the first message is computed by calling the update \texttt{ruleVBGaussianMixtureZCat} on a data entry (\texttt{data[:y][8]}) and a pre-initialized dictionary of marginal beliefs. Finally, an updated dictionary of marginal beliefs is returned.}
\label{lst:algo}
\end{figure}
\begin{figure}[ht]
\begin{minted}[baselinestretch=1.2,bgcolor=LightGray,fontsize=\footnotesize]{julia}
function freeEnergy(data::Dict, marginals::Dict)
    F = 0.0
    F += averageEnergy(Dirichlet, marginals[:T], ProbabilityDistribution(MatrixVariate, PointMass,
        m=[1.0 1.0 1.0; 1.0 1.0 1.0; 1.0 1.0 1.0]))
    ...
    F += averageEnergy(GaussianMixture, ProbabilityDistribution(Multivariate, PointMass, m=data[:y][1]),
        marginals[:x_1], marginals[:m1], marginals[:W1], marginals[:m2], marginals[:W2], marginals[:m3],
        marginals[:W3])
    ...
    F -= differentialEntropy(marginals[:T])
    F -= differentialEntropy(marginals[:m1])
    ...
    return F
end
\end{minted}
\caption{Segment of automatically generated code for evaluating the variational free energy as stored in \texttt{algo\_F} (Fig.~\ref{lst:schedule}). The free energy is computed by summing and subtracting (local) energy and entropy terms (see also Eq.~\ref{eq:FE-energy-entropy}).}
\label{lst:algo_F}
\end{figure}

\begin{figure}[h]
    \center
    \begin{subfigure}{.33\textwidth}
        \center
        \resizebox{\textwidth}{!}{\includegraphics{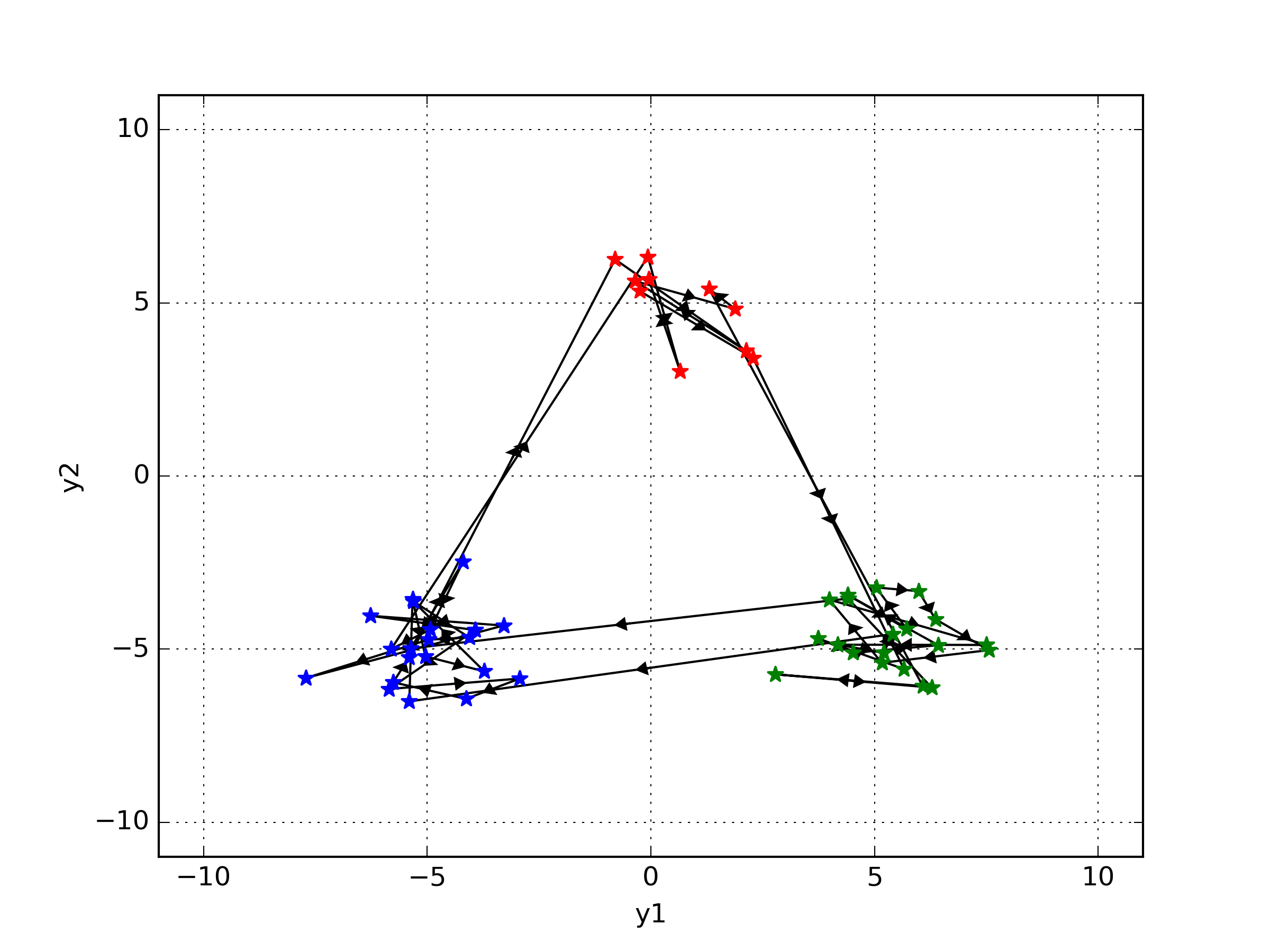}}
    \end{subfigure}
    \begin{subfigure}{.33\textwidth}
        \center
        \resizebox{\textwidth}{!}{\includegraphics{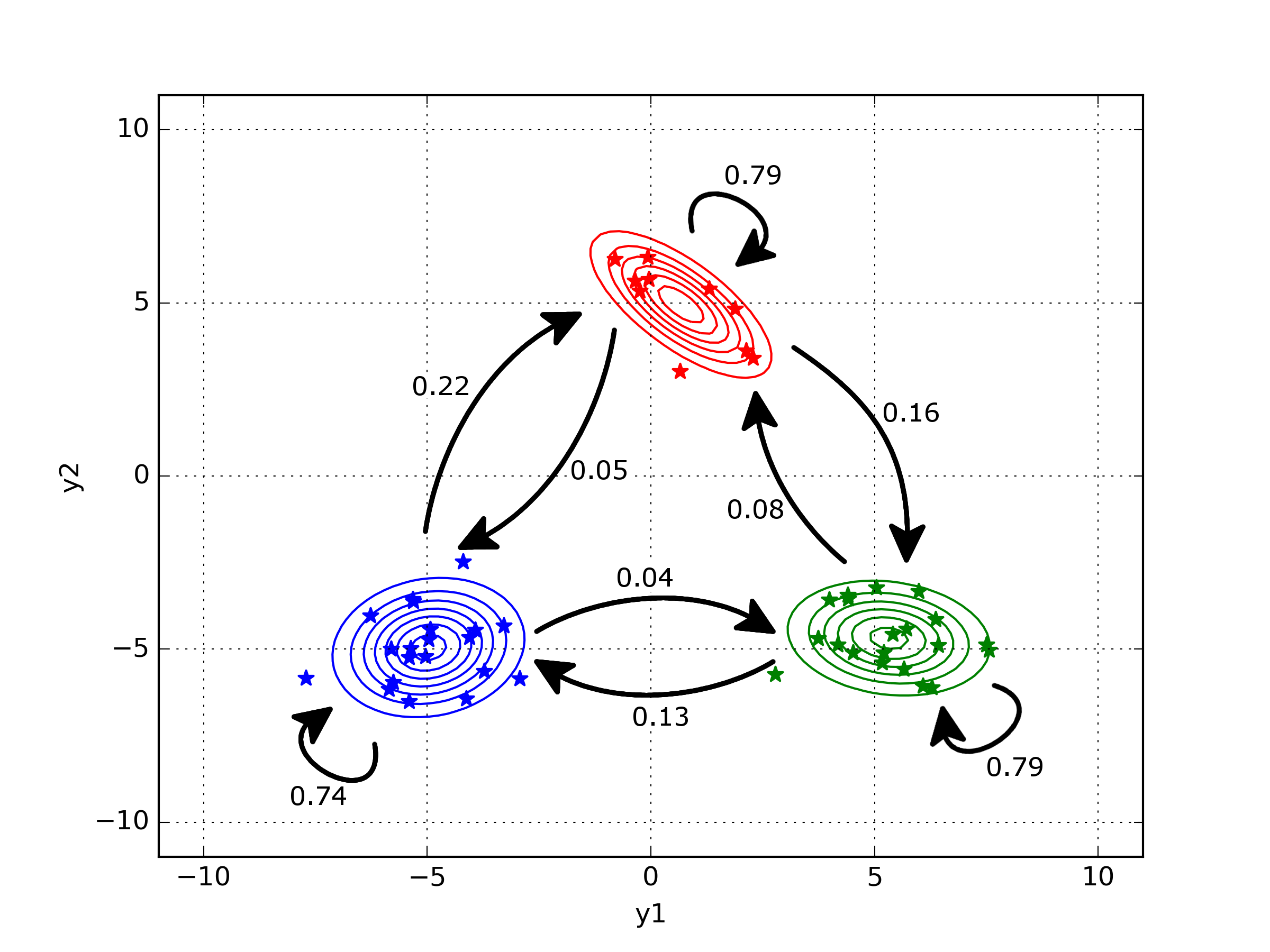}}
    \end{subfigure}
    \begin{subfigure}{.33\textwidth}
        \center
        \resizebox{\textwidth}{!}{\includegraphics{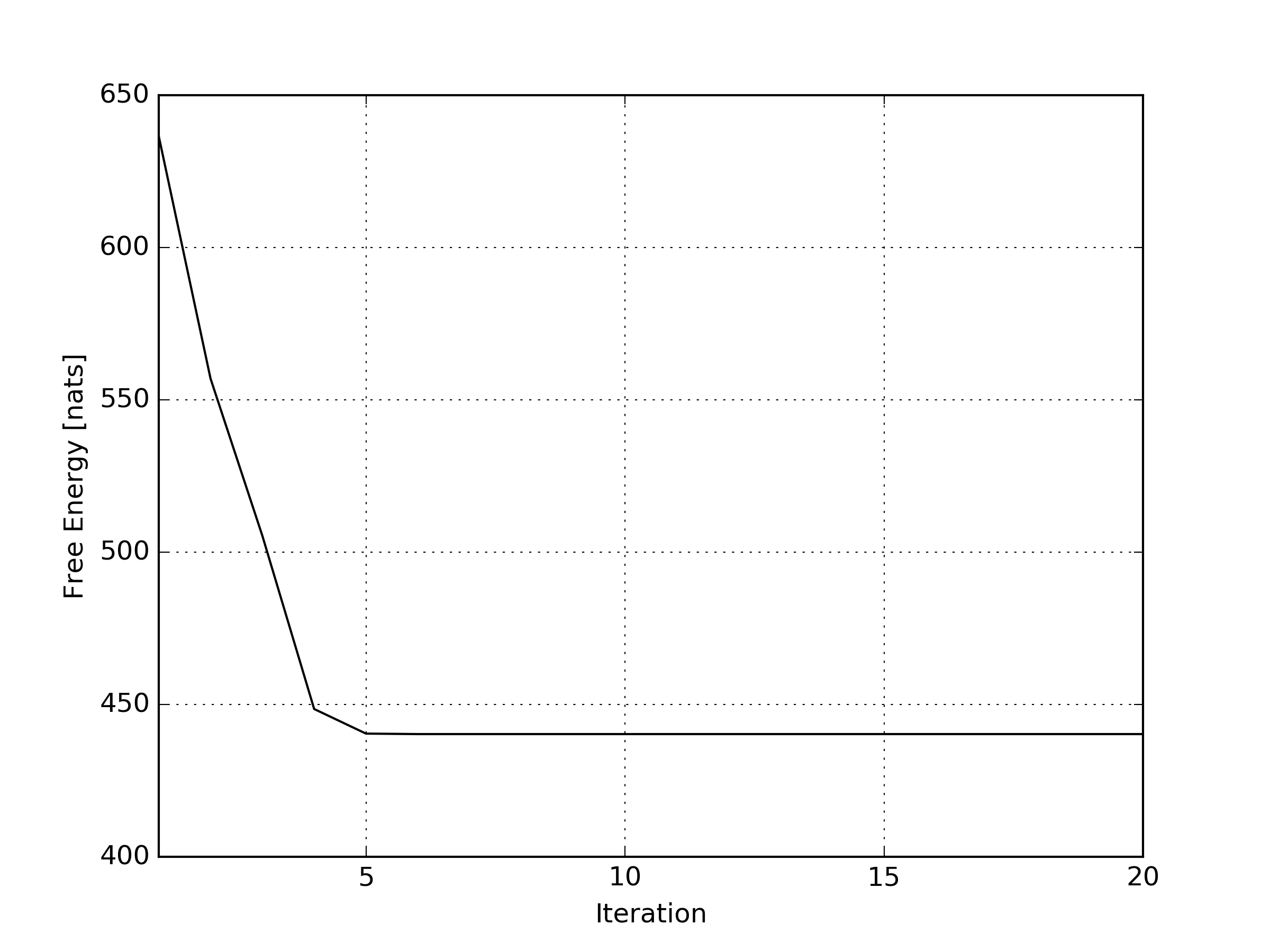}}
    \end{subfigure}

    \caption{Left: synthetic two-dimensional data set generated by sampling a sequence of 50 time steps from a HMGM model with $K=3$ components. Colors correspond to the latent hidden state of the Markov model, arrows indicate the sequence of observations. The parameters of the generative model are chosen to only allow `clockwise' state transitions.
    Middle: visualization of the inferred model parameters (Gaussian mixture model and state transition probabilities) after convergence of the variational Bayes algorithm.
    Right: evolution of the variational free energy during execution of the inference algorithm.}
    \label{fig:dsco_est}
\end{figure}

Fig.~\ref{lst:infer} shows example code for executing the generated inference algorithm. This involves setting the initial recognition distributions, collecting observed variables and repeatedly executing the algorithm until convergence.
To test the algorithm, we apply it on a synthetic data set sampled from the generative model with fixed parameters. Fig.~\ref{fig:dsco_est} (left) shows the data set, which contains 50 two-dimensional observations drawn from a HMGM model with $K=3$ components. Fig.~\ref{fig:dsco_est} (middle) visualizes the inferred model parameters (means of the approximate posterior distributions) after convergence of the inference algorithm. The solution correctly identifies all Gaussian mixture components in the observation model, and finds appropriate state transition probabilities for the hidden Markov model that governs the component switches. The variational free energy converges to a local minimum after roughly 5 iterations of the VMP algorithm, as shown by Fig.~\ref{fig:dsco_est} (right).
\begin{figure}[ht]
\begin{minted}[baselinestretch=1.2,bgcolor=LightGray,fontsize=\footnotesize]{julia}
# Load algorithms (algo and algo_F are strings containing the source code)
eval(parse(algo))
eval(parse(algo_F))

# Initial recognition distributions
marginals = Dict{Symbol, ProbabilityDistribution}(
    :T  => vague(Dirichlet, (3,3)),
    :m1 => ProbabilityDistribution(Multivariate, GaussianMeanVariance, m=[0.0, 1.0], v=100.0*diageye(2)),
    :W1 => ProbabilityDistribution(Wishart, v=10.0*diageye(2), nu=2.0),
    ...
)

# Initialize data
data = Dict(:y => y_data)
n_its = 20

# Execute algorithm by iteratively calling automatically generated functions
F = Vector{Float64}(n_its)
for i = 1:n_its
    stepX!(data, marginals)
    stepW1!(data, marginals)
    ...
    stepM1!(data, marginals)
    ...
    stepT!(data, marginals)

    F[i] = freeEnergy(data, marginals)
end
\end{minted}
\caption{Julia code snippet for executing the VMP algorithm generated by ForneyLab. The first two lines of code parse and evaluate the automatically generated algorithm code from Fig.~\ref{lst:schedule}. This imports the recognition factor-specific \texttt{step!} functions (Fig.~\ref{lst:algo}) and the \texttt{freeEnergy} function (Fig.~\ref{lst:algo_F}) into the current scope. The \texttt{step!} functions can then be iterated to update (in-place) a pre-initialized dictionary of marginal beliefs. After each iteration the free energy can be computed and inspected for convergence.}
\label{lst:infer}
\end{figure}

The automatically generated algorithm for full Bayesian inference in the HMGM model is an illustrative example of how the message passing paradigm can help to produce efficient algorithms without requiring expert knowledge about the underlying methods. While it is possible to manually derive a variational inference algorithm for the model at hand, it involves long and tedious (model-dependent) derivations. Using the divide-and-conquer approach of message passing, we are able to reduce this task to deriving update rules for individual factor nodes, which is much simpler. Moreover, these update rules can be reused to perform inference in other models involving the same factors. Note that while black-box inference methods like automatic differentiation variational inference (ADVI) \cite{kucukelbir_automatic_2017} require no manual derivations at all, they are generally much slower than (message passing) algorithms that leverage analytic solutions. Moreover, additional tricks are required to make ADVI possible in models involving discrete latent variables (see e.g. \cite{tucker_rebar:_2017}). In Sec.~\ref{sec:efficient} we compare message passing-based inference to black-box inference methods in terms of performance.

\FloatBarrier
\section{Boosting the algorithm design loop by automated inference in factor graphs}
\label{sec:effective_flexible}

In the search process for an effective algorithm, it is important that we can quickly specify updates to model proposals and compare the relative performance of these proposals. Here, we illustrate how the modularity of the FFG framework allows for flexible construction of custom models and algorithms. In Sec.~\ref{sec:lgm} we model a time series with a linear Gaussian model. This model is improved in Sec.~\ref{sec:adaptation}, where we exemplify how the model can be adapted by explicitly incorporating nonlinearities. With ForneyLab, this adaptation evaluates to adding one extra line of code to the model specification. In Sec.~\ref{sec:combining}, we showcase the flexibility with respect to algorithm specification by constructing a custom algorithm that combines VMP with expectation propagation. We will see that the combined benefits of both algorithms leads to improved performance over pure VMP in our example. Finally, Sec.~\ref{sec:composability} underlines the hierarchical nature of the FFG framework by considering \emph{composite} nodes as modular building blocks for model construction. This construct allows for building complex hierarchical structures, and improved inference by leveraging external tools and custom updates.

\subsection{Time series modeling with a linear Gaussian SSM}
\label{sec:lgm}

In this section we provide an example of a linear Gaussian SSM that we will later adapt (Sec.~\ref{sec:adaptation}) to improve the model fit to a simulated data set. Consider a time series data set consisting of hourly temperature measurements over a period of two days generated by the following process:
\begin{subequations}
\begin{align}
    \hat{\vect{w}}_t &\sim \N{\vect{0}, \hat{\matr{W}}^{-1}} \label{eq:lgm_gp_a}\\
    \hat{\vect{x}}_t &= \matr{A} \hat{\vect{x}}_{t-1} + \hat{\vect{w}}_t \label{eq:lgm_gp_b}\\
    \hat{v}_t &\sim \N{0, \hat{u}^{-1}}\\
    \hat{y}_t &= \log \left(1 + \exp{\vect{b}^{\T}\,\hat{\vect{x}}_t}\right) + \hat{v}_t\,. \label{eq:lgm_gp_d}
\end{align}
\end{subequations}
In order to make things interesting, we assumed that our thermometer has a nonlinear response curve, given by the softplus function $g(x) = \log \left(1 + \exp{x}\right)$, which truncates negative temperatures. In order to introduce a daily periodicity in the temperatures, the hidden state represents a phasor $\phi_t \in \mathbb{C}$, of which the real and imaginary component are respectively encoded by the entries in the hidden states $\hat{\vect{x}}_t \in \mathbb{R}^2$. Two days ($T = 48$) of data are generated from an initial state $\hat{\vect{x}}_0 = (5, 0)^{\T}$ by recursive application of Eq.~\ref{eq:lgm_gp_a}--\ref{eq:lgm_gp_d}. The state transition matrix $\matr{A}$ represents a rotation with angular frequency $\pi/12$ and $\hat{\matr{W}}$ is a diagonal precision matrix with $\hat{W}_{jj}=50, \hat{W}_{i\neq j}=0$. The observation is generated from the first (real) state vector component by the selection vector $\vect{b} = (1, 0)^{\T}$. Finally, $\hat{u} = 50$ represents the precision of the Gaussian observation noise. The resulting observations and hidden states are shown in Fig.~\ref{fig:glm_est}.

In order to model this time series we postulate a linear Gaussian dynamic model as defined by
\begin{subequations}
\begin{align}
    p(\vect{x}_t \,|\, \vect{x}_{t-1}, \matr{W}) &= \mathcal{N} \left( \vect{x}_t \,\middle|\, \matr{A}\vect{x}_{t-1}, \matr{W}^{-1} \right) \label{eq:lgm_transition}\\
    p(y_t \,|\, \vect{x}_t, u) &= \mathcal{N} \left( y_t \,\middle|\, \vect{b}^{\T}\vect{x}_t, u^{-1} \right) \label{eq:lgm_observation}\,,
\end{align}
\label{eq:lgm}
\end{subequations}
with ``vague'' (virtually uninformative) priors on the initial state and precisions. Because this generative model for the data lacks the softplus nonlinearity that was used in generating the data, there exists a deliberate mismatch between the generative process and our proposed generative model. In Sec.~\ref{sec:adaptation} we will alleviate this mismatch by explicitly modeling the nonlinearity as well.

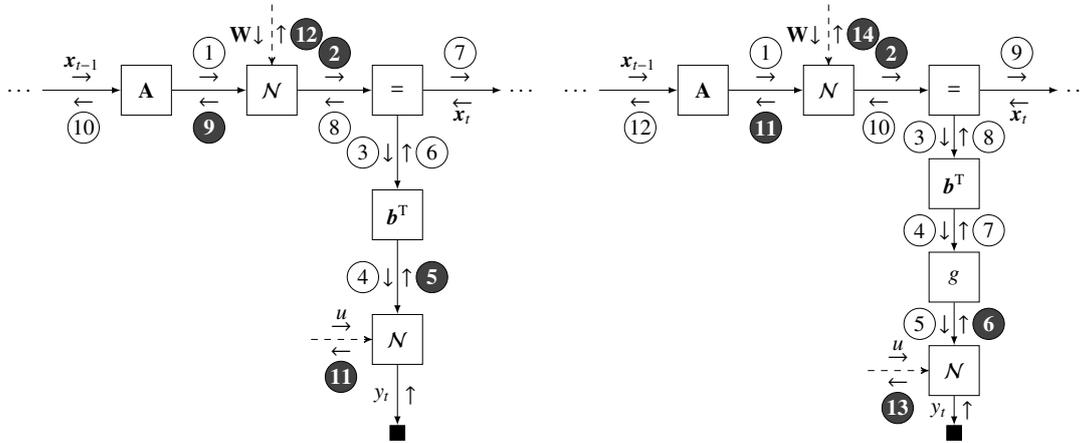
\begin{figure}[h]
    \center
    \begin{subfigure}{.44\textwidth}
        \center
        \resizebox{\textwidth}{!}{\begin{tikzpicture}
    [node distance=20mm,auto,>=stealth']


    \node[box] (A) {$\matr{A}$};
    \node[left of=A] (x_t_min_dots) {$\dots$};
    \node[box, right of=A] (N_x) {$\mathcal{N}$};
    \node[above of=N_x, node distance=15mm] (W) {};
    \node[box, right of=N_x] (x_eq) {$=$};
    \node[right of=x_eq] (x_t_dots) {$\dots$};

    \node[box, below of=x_eq] (b) {$\vect{b}^{\T}$};
    \node[box, below of=b] (N_y) {$\mathcal{N}$};
    \node[left of=N_y, node distance=15mm] (u) {};
    \node[clamped, below of=N_y, node distance=15mm] (y_t) {};

    \path[line] (x_t_min_dots) edge[->] node[anchor=south, yshift=2mm]{$\vect{x}_{t-1}$} node[anchor=south]{$\rightarrow$} (A);
    \path[line] (A) edge[->] (N_x);
    \path[dashed] (W) edge[->] node[anchor=east, xshift=-2mm]{$\matr{W}$} node[anchor=east]{$\downarrow$} (N_x);
    \path[line] (N_x) edge[->] (x_eq);
    \path[line] (x_eq) edge[->] node[anchor=north, yshift=-2mm]{$\vect{x}_t$} node[anchor=north]{$\leftarrow$} (x_t_dots);

    \path[line] (x_eq) edge[->] (b);
    \path[line] (b) edge[->] (N_y);
    \path[dashed] (u) edge[->] node[anchor=south, yshift=2mm]{$u$} node[anchor=south]{$\rightarrow$} (N_y);
    \path[line] (N_y) edge[->] node[anchor=east]{$y_t$} node[anchor=west]{$\uparrow$} (y_t);

    \msg{up}{right}{A}{N_x}{0.5}{1};
    \darkmsg{up}{right}{N_x}{x_eq}{0.5}{2};
    \msg{left}{down}{x_eq}{b}{0.5}{3};
    \msg{left}{down}{b}{N_y}{0.5}{4};
    \darkmsg{right}{up}{N_y}{b}{0.5}{5};
    \msg{right}{up}{b}{x_eq}{0.5}{6};
    \msg{up}{right}{x_eq}{x_t_dots}{0.5}{7};

    \msg{down}{left}{N_x}{x_eq}{0.5}{8};
    \darkmsg{down}{left}{A}{N_x}{0.5}{9};
    \msg{down}{left}{x_t_min_dots}{A}{0.5}{10};
    \darkmsg{down}{left}{u}{N_y}{0.4}{11};
    \darkmsg{right}{up}{N_x}{W}{0.6}{12};
\end{tikzpicture}}
    \end{subfigure}
    \begin{subfigure}{.44\textwidth}
        \center
        \resizebox{\textwidth}{!}{\begin{tikzpicture}
    [node distance=20mm,auto,>=stealth']


    \node[box] (A) {$\matr{A}$};
    \node[left of=A] (x_t_min_dots) {$\dots$};
    \node[box, right of=A] (N_x) {$\mathcal{N}$};
    \node[above of=N_x, node distance=15mm] (W) {};
    \node[box, right of=N_x] (x_eq) {$=$};
    \node[right of=x_eq] (x_t_dots) {$\dots$};

    \node[box, below of=x_eq, node distance=15mm] (b) {$\vect{b}^{\T}$};
    \node[box, below of=b, node distance=15mm] (g) {$g$};
    \node[box, below of=g, node distance=15mm] (N_y) {$\mathcal{N}$};
    \node[left of=N_y, node distance=15mm] (u) {};
    \node[clamped, below of=N_y, node distance=10mm] (y_t) {};

    \path[line] (x_t_min_dots) edge[->] node[anchor=south, yshift=2mm]{$\vect{x}_{t-1}$} node[anchor=south]{$\rightarrow$} (A);
    \path[line] (A) edge[->] (N_x);
    \path[dashed] (W) edge[->] node[anchor=east, xshift=-2mm]{$\matr{W}$} node[anchor=east]{$\downarrow$} (N_x);
    \path[line] (N_x) edge[->] (x_eq);
    \path[line] (x_eq) edge[->] node[anchor=north, yshift=-2mm]{$\vect{x}_t$} node[anchor=north]{$\leftarrow$} (x_t_dots);

    \path[line] (x_eq) edge[->] (b);
    \path[line] (b) edge[->] (g);
    \path[line] (g) edge[->] (N_y);
    \path[dashed] (u) edge[->] node[anchor=south, yshift=2mm]{$u$} node[anchor=south]{$\rightarrow$} (N_y);
    \path[line] (N_y) edge[->] node[anchor=east]{$y_t$} node[anchor=west]{$\uparrow$} (y_t);

    \msg{up}{right}{A}{N_x}{0.5}{1};
    \darkmsg{up}{right}{N_x}{x_eq}{0.5}{2};
    \msg{left}{down}{x_eq}{b}{0.5}{3};
    \msg{left}{down}{b}{g}{0.5}{4};
    \msg{left}{down}{g}{N_y}{0.5}{5};
    \darkmsg{right}{up}{N_y}{g}{0.5}{6};
    \msg{right}{up}{g}{b}{0.5}{7};
    \msg{right}{up}{b}{x_eq}{0.5}{8};
    \msg{up}{right}{x_eq}{x_t_dots}{0.5}{9};

    \msg{down}{left}{N_x}{x_eq}{0.4}{10};
    \darkmsg{down}{left}{A}{N_x}{0.5}{11};
    \msg{down}{left}{x_t_min_dots}{A}{0.5}{12};
    \darkmsg{down}{left}{u}{N_y}{0.4}{13};
    \darkmsg{right}{up}{N_x}{W}{0.6}{14};
\end{tikzpicture}}
    \end{subfigure}
    \caption{Message passing schedule (left) for estimation on a linear Gaussian state-space model (see Eq.~\ref{eq:lgm}; priors not drawn). Black-labeled messages are computed through the variational message passing update rule from \cite{dauwels_variational_2007}. The right figure shows the message passing schedule for the model with the extended (nonlinear) observation model introduced in Eq.~\ref{eq:nlgm_observation}.}
    \label{fig:lst}
\end{figure}

Assume that we are interested in estimating a posterior belief for the hidden state sequence ($\vect{x}$), and the transition and observation noise precisions ($\matr{W}$ and $u$) from a given data set ($y=\hat{y}$). In other words, we are interested in evaluating the inference task
\begin{align*}
    p(\vect{x}, \matr{W}, u\,|\, y) &= \frac{p(y, \vect{x}, \matr{W}, u)}{\idotsint p(y, \vect{x}, \matr{W}, u) \d{\vect{x}} \d{\matr{W}} \d{u}}\,.
\end{align*}
In order to evaluate this inference task, we perform approximate inference by variational message passing \cite{dauwels_variational_2007} and choose the recognition distribution factorization
\begin{align*}
    q(\vect{x}, \matr{W}, u) &= q(\vect{x}) \, q(\matr{W}) \, q(u)\,,
\end{align*}
which imposes a \emph{structured} factorization of $q$ by assuming a single joint recognition factor for the full state sequence $\vect{x} = (\vect{x}_0,\vect{x}_1,\dots,\vect{x}_T)$, and a single joint recognition factor for all entries of the precision matrix $\matr{W}$. The resulting message passing schedule for the current problem is illustrated in Fig.~\ref{fig:lst} (left). First, messages $\circled{1}$ and $\darkcircled{2}$ estimate a current state from the previous state, and messages $\circled{3}$ and $\circled{4}$ propagate predictions towards the current observation. Next, the current state estimate is corrected by messages $\darkcircled{5}$, $\circled{6}$ and $\circled{7}$, which account for evidence contained by the current observation. Propagating forward, these messages execute a forward (filtering) pass over the full state sequence (i.e., going forward from $t=0$ to $t=T$). This is followed by a backward (smoothing) pass comprising the messages $\circled{8}$, $\darkcircled{9}$ and $\circled{10}$ that runs from $t=T$ backwards to $t=0$. The smoothing pass improves the state estimates through evidence (observations) from future time steps. In a real-time processing scenario, we may need to skip the smoothing pass. Finally, given the updated state estimates, messages $\darkcircled{11}$ and $\darkcircled{12}$ update the estimates for the precision parameters. This message passing sequence constitutes one iteration of the estimation process. The performance of the estimation process can be improved by repeating the sequence over multiple iterations, where the resulting estimates of an iteration are taken as initial estimates (priors) for the next iteration.

The simulation results for the state estimate after $100$ iterations is shown in Fig.~\ref{fig:glm_est} (left). It can be seen that the estimated state faithfully follows the observed data. Posterior precisions are $q(u) = \textrm{Gam}\left( u \,|\, a=25, b=0.36 \right)$, and $q(\matr{W}) = \mathcal{W}\left(\matr{W} \,\middle|\, \matr{V}=\begin{pmatrix}3.87 & 0.51 \\ 0.51 & 0.09\end{pmatrix}, \nu=50 \right)$ (where $\mathcal{W}$ denotes a Wishart distribution). Fig.~\ref{fig:glm_est} (right) shows the free energy as a function of number of iterations. 

\subsection{Trying an alternative model with a nonlinear likelihood}
\label{sec:adaptation}

We now try to improve the algorithm performance of the SSM by postulating an alternative observation model. Specifically, we will explicitly account for the nonlinear corruption of the measurements. Without an automated-inference toolbox, any model adaptation would require a possibly tedious manual re-derivation of the inference update equations. In the factor graph framework, the probabilistic model can be readily extended, and the adjusted VMP algorithm can be automatically derived. With ForneyLab, this extension evaluates to a single extra line in the model definition. With the softplus nonlinearity $g(\cdot)$ included, the observation model becomes
\begin{align}
    p(y_t \,|\, \vect{x}_t, u) &= \mathcal{N}\left(y_t \,\middle|\, g(\vect{b}^{\T}\vect{x}_t), u^{-1}\right) \label{eq:nlgm_observation}\,.
\end{align}
The message passing schedule for the nonlinear model is drawn in Fig.~\ref{fig:lst} (right). Exact computation of message \circled{5} is complicated by the nonlinearity introduced by the $g$ node. However, message passing allows us to retain conjugacy by computing messages through local approximations. Therefore, we linearize $g(\cdot)$ around the mean of the inbound message \circled{4}, and compute the outbound message from the approximated node function. The same procedure is applied to message \circled{7}. While this local approximation introduces an error in the resulting message computations, we will see that it works well in practice.

\begin{figure}[ht]
    \center
    \begin{subfigure}{.49\textwidth}
        \center
        \resizebox{\textwidth}{!}{\includegraphics{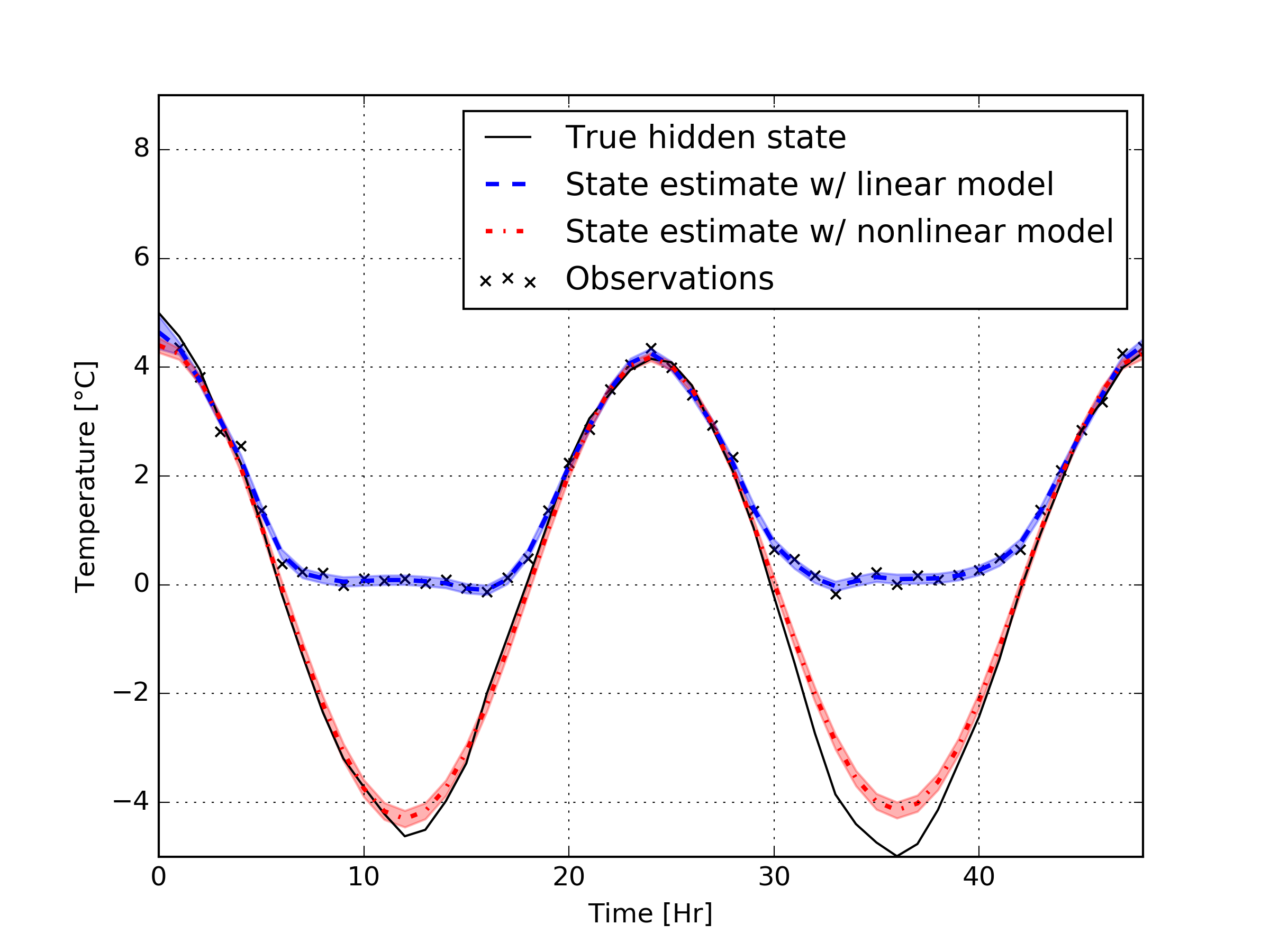}}
    \end{subfigure}
    \begin{subfigure}{.49\textwidth}
        \center
        \resizebox{\textwidth}{!}{\includegraphics{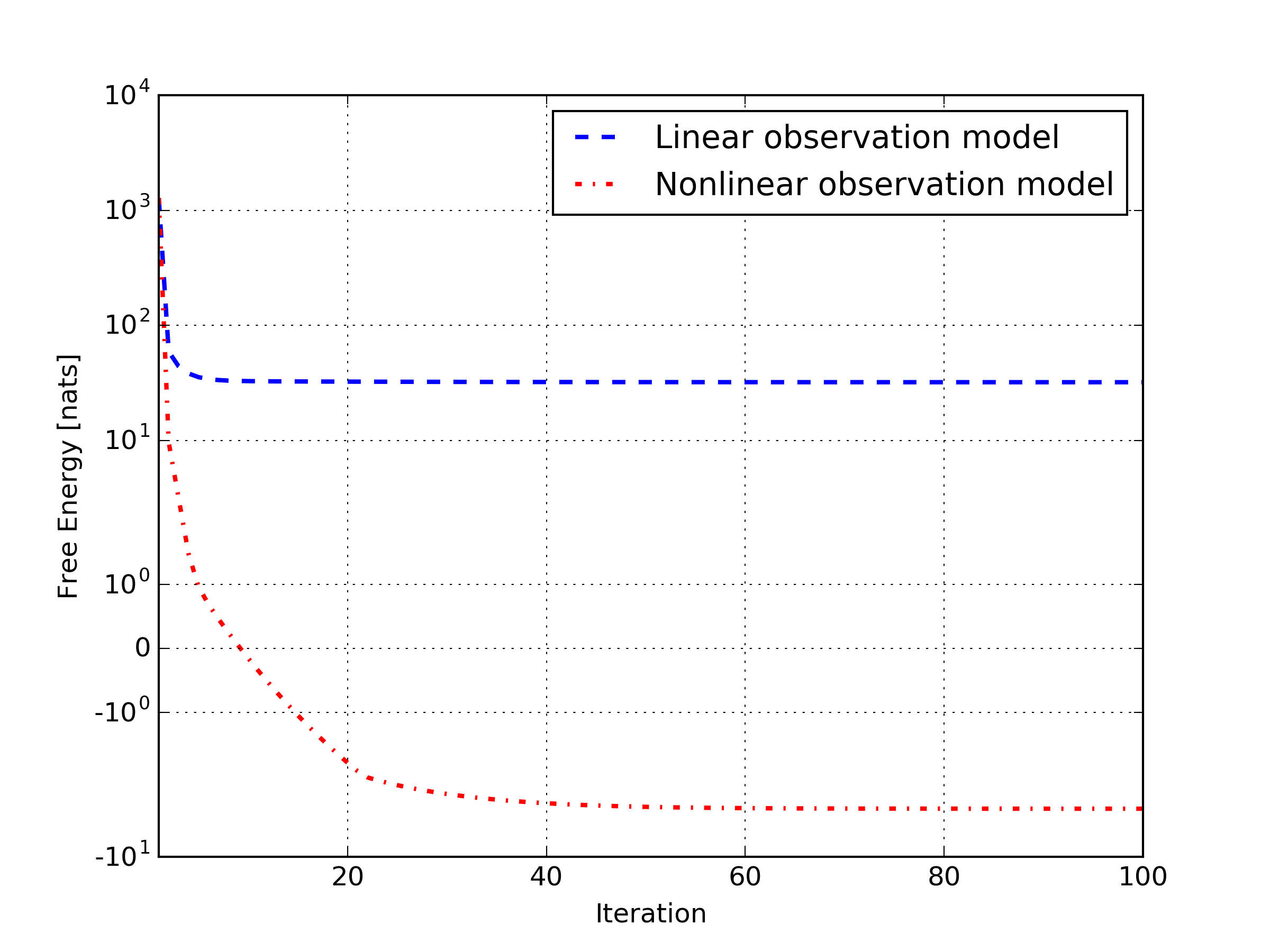}}
    \end{subfigure}
    \caption{Inference results for estimation (left) of linear and nonlinear Gaussian state-space models on a toy data set. The right figure shows the convergence of the free energy for both cases.}
    \label{fig:glm_est}
\end{figure}

Estimation of the alternative model with ForneyLab yields the state (first component) estimation result of Fig.~\ref{fig:glm_est} (left), precision estimates $q(u) = \mathrm{Gam}\left( u \,|\, a=25, b=0.50\right)$ and $q(\matr{W}) = \mathcal{W}\left(\matr{W} \,\middle|\, \matr{V}=\begin{pmatrix}5.65 & 0.00 \\ 0.00 & 5.62\end{pmatrix}, \nu=50 \right)$, and free energy estimate in Fig.~\ref{fig:glm_est} (right). The final difference in free energy between the linear and nonlinear model evaluates to $156$ [dB], which rules overwhelmingly in favor of the nonlinear model.

This section intended to exemplify the ease with which an alternative model proposal can be scored on performance. By simply replacing or adding factor nodes, alternative models can be specified and ForneyLab automatically delivers code for inference algorithms, including code for performance evaluation. This eliminates the need for tedious manual derivations, and opens up the possibility of fast iterative search for the best model fit to the data \cite{blei_build_2014}. Furthermore, we illustrated how message passing allows for local approximations to difficult messages resulting from nonlinearities in the model.

\FloatBarrier
\subsection{Message passing on FFGs as a platform for combining inference algorithms}
\label{sec:combining}

In this section we exemplify how message passing with ForneyLab combines VMP with expectation propagation (EP) for estimating a SSM with a binary observation model. Similar models are often used in the context of perception and decision making, e.g. \cite{mathys_uncertainty_2014, bitzer_perceptual_2014}. Here we exemplify how a hybrid VMP-EP algorithm leads to improved estimation results over full VMP on a binary model example.

In this example, we use the same hidden state data generating process as in Sec.~\ref{sec:lgm}. We generate a hidden state sequence by Eq.~\ref{eq:lgm_gp_a} and \ref{eq:lgm_gp_b} with initial state $\hat{x}_0 = (1, 0)^{\T}$. Then, in contrast to Sec.~\ref{sec:lgm}, we draw $T = 96$ binary observations $\hat{y}_t \in \{1, -1\}$ (true, false), where the probability of the outcomes is determined by the fluctuating continuous hidden state, as
\begin{subequations}
\begin{align}
    Pr(\hat{y}_t = 1) = \Phi(\vect{b}^{\T} \hat{\vect{x}}_t)\,,
\end{align}
\end{subequations}
where $\Phi(\cdot)$ is the standard Gaussian cumulative density function.

For the generative model we assume Eq.~\ref{eq:lgm_transition} for the state transition model, and define the observation model as
\begin{align}
    p(y_t \,|\, \vect{x}_t) &= \Phi \left( y_t \cdot \vect{b}^{\T} \vect{x}_t \right) \,. \label{eq:sigmoid_observation}
\end{align}
The full generative model is obtained by substituting Eqs.~\ref{eq:sigmoid_observation} and \ref{eq:lgm_transition} in Eq.~\ref{eq:ssm_gm}, and choosing vague priors.

Assume that we are interested in inferring a posterior belief over the hidden state sequence ($\vect{x}$) and the transition precision ($\matr{W}$) from the observed discrete data set ($y = \hat{y}$). As before, we choose a structured recognition distribution factorization, given by
\begin{align}
    q(\vect{x}, \matr{W}) &= q(\vect{x}) \, q(\matr{W})\,. \label{eq:q_vmp_ep}
\end{align}

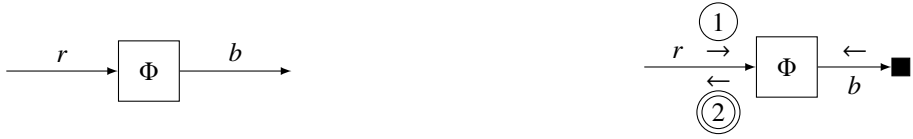
\begin{figure}[h]
    \begin{subfigure}{.49\textwidth}
        \center
        \begin{tikzpicture}
    [node distance=20mm,auto,>=stealth']


    \node[box] (sig) {$\Phi$};

    \node[left of=sig] (r) {};
    \node[right of=sig] (b) {};

    \path[line] (r) edge[->] node[anchor=south]{$r$} (sig);
    \path[line] (sig) edge[->] node[anchor=south]{$b$} (b);
\end{tikzpicture}
    \end{subfigure}
    \begin{subfigure}{.49\textwidth}
        \center
        \begin{tikzpicture}
    [node distance=20mm,auto,>=stealth']


    \node[box] (sig) {$\Phi$};

    \node[left of=sig] (r) {};
    \node[clamped, right of=sig, node distance=15mm] (b) {};

    \path[line] (r) edge[->] node[anchor=south, pos=0.3]{$r$} (sig);
    \path[line] (sig) edge[->] node[anchor=north]{$b$} node[anchor=south]{$\leftarrow$} (b);

    \msg{up}{right}{r}{sig}{0.55}{1};
    \bwmsg{down}{left}{r}{sig}{0.55}{2};
\end{tikzpicture}
    \end{subfigure}
    \caption{FFG representation (left) and message passing schedule with observed datum (right) for a binary node. Here, the double circled message represents an expectation propagation message \cite{cox_robust_nodate}.}
    \label{fig:sigmoid}
\end{figure}

However, in formulating the message passing algorithm, we immediately run into trouble with the factor $f_{\Phi}(b, r) = \Phi(b\cdot r)$, linking a binary variable $b \in \{1, -1\}$ to a real variable $r \in \mathbb{R}$. We first zoom in on this binary factor node, which is drawn in Fig.~\ref{fig:sigmoid} (left).

Naively, we might attempt to compute the backward message (indicated by a left overhead arrow) for an observation $\hat{b} = 1$ as
\begin{align*}
    \overleftarrow{\mu}(r) = \sum_{b\in\{1, -1\}} f_{\Phi}(b, r)\, \delta(b - \hat{b}) = \sum_{b\in\{1, -1\}} \Phi(b\cdot r)\, \delta(b - 1) = \Phi(r)\,.
\end{align*}
However, this message breaks conjugacy and leads to increasingly complex messages when propagated further into the model. In an effort to obtain conjugate updates, some proposals mend the variational message updates \cite{knowles_non-conjugate_2011, nolan_accurate_2017}, while others augment the generative model \cite{albert_bayesian_1993}. In this section, we propose an alternative approach that takes full advantage of the modularity of the FFG framework. The intrinsic modularity allows for selecting different Bayesian approximation methods at each node-edge interface in the generative model. Thus, the FFG framework allows not only to change local model assumptions, but also to change local inference methods.

\begin{figure}[h]
    \center
    \begin{subfigure}{.44\textwidth}
        \center
        \resizebox{\textwidth}{!}{\begin{tikzpicture}
    [node distance=20mm,auto,>=stealth']


    \node[box] (A) {$\matr{A}$};
    \node[left of=A] (x_t_min_dots) {$\dots$};
    \node[box, right of=A] (N_x) {$\mathcal{N}$};
    \node[above of=N_x, node distance=15mm] (W) {};
    \node[box, right of=N_x] (x_eq) {$=$};
    \node[right of=x_eq] (x_t_dots) {$\dots$};

    \node[box, below of=x_eq] (b) {$\vect{b}^{\T}$};
    \node[box, below of=b] (sigmoid) {$\Phi$};
    \node[clamped, below of=sigmoid, node distance=15mm] (y_t) {};


    \path[line] (x_t_min_dots) edge[->] node[anchor=south, yshift=2mm]{$\vect{x}_{t-1}$} node[anchor=south]{$\rightarrow$} (A);
    \path[line] (A) edge[->] (N_x);
    \path[dashed] (W) edge[->] node[anchor=east, xshift=-2mm]{$\matr{W}$} node[anchor=east]{$\downarrow$} (N_x);
    \path[line] (N_x) edge[->] (x_eq);
    \path[line] (x_eq) edge[->] node[anchor=north, yshift=-2mm]{$\vect{x}_t$} node[anchor=north]{$\leftarrow$} (x_t_dots);

    \path[line] (x_eq) edge[->] (b);
    \path[line] (b) edge[->] (sigmoid);
    \path[line] (sigmoid) edge[->] node[anchor=east]{$y_t$} node[anchor=west]{$\uparrow$} (y_t);

    \msg{up}{right}{A}{N_x}{0.5}{1};
    \darkmsg{up}{right}{N_x}{x_eq}{0.5}{2};
    \msg{left}{down}{x_eq}{b}{0.5}{3};
    \msg{left}{down}{b}{sigmoid}{0.5}{4};
    \bwmsg{right}{up}{sigmoid}{b}{0.5}{5};
    \msg{right}{up}{b}{x_eq}{0.5}{6};
    \msg{up}{right}{x_eq}{x_t_dots}{0.5}{7};

    \msg{down}{left}{N_x}{x_eq}{0.5}{8};
    \darkmsg{down}{left}{A}{N_x}{0.5}{9};
    \msg{down}{left}{x_t_min_dots}{A}{0.5}{10};
    \darkmsg{right}{up}{N_x}{W}{0.6}{11};

\end{tikzpicture}}
    \end{subfigure}
    \begin{subfigure}{.54\textwidth}
        \center
        \resizebox{\textwidth}{!}{\includegraphics{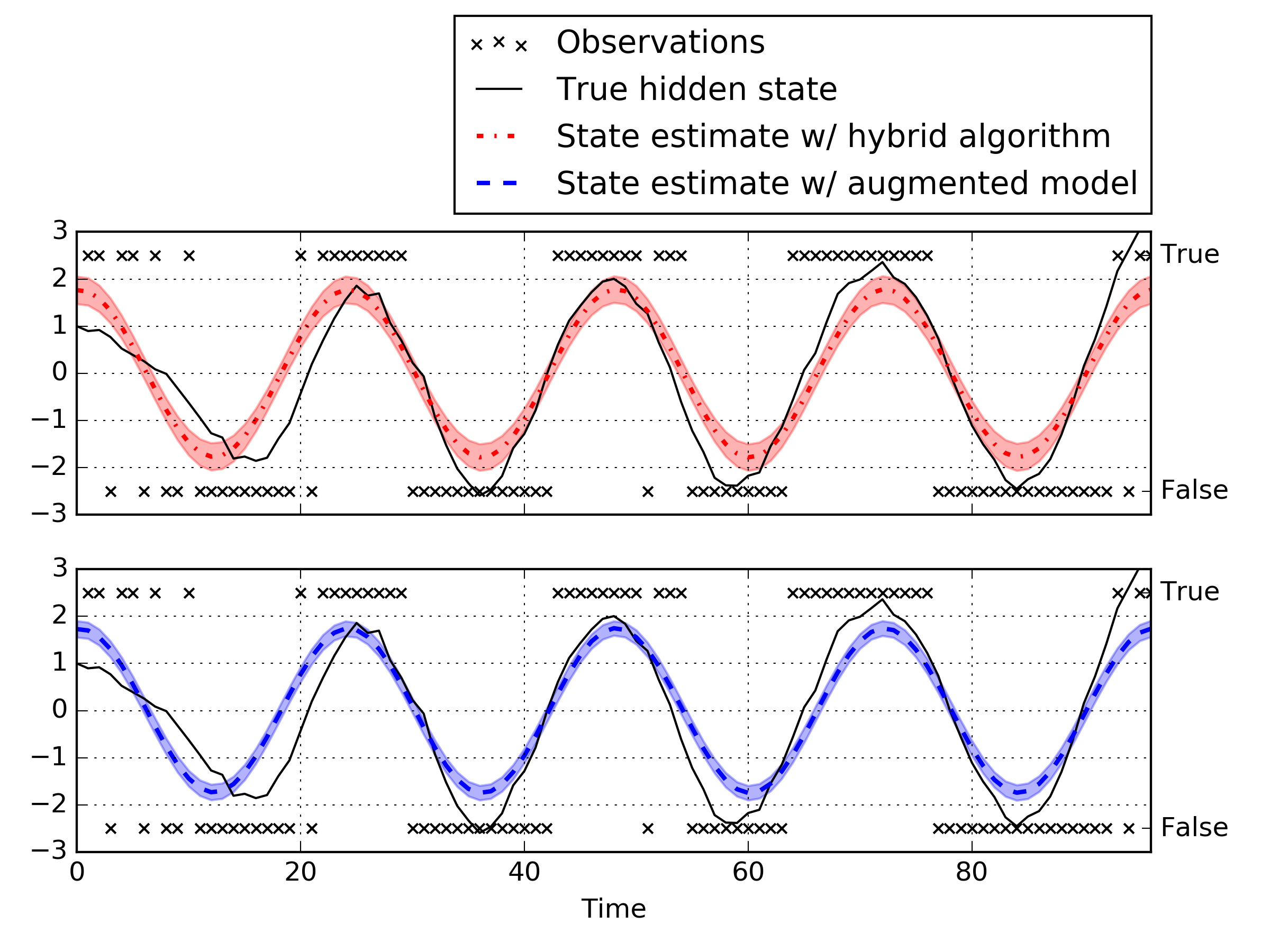}}
    \end{subfigure}
    \caption{VMP-EP message passing schedule (left) for estimation on a linear Gaussian SSM with sigmoid observation model, together with a comparison of the hidden state estimates for VMP-EP (top right) and VMP with model augmentation (bottom right). The doubly circled message is computed by the EP update rule \cite{cox_robust_nodate}.}
    \label{fig:csdo}
\end{figure}
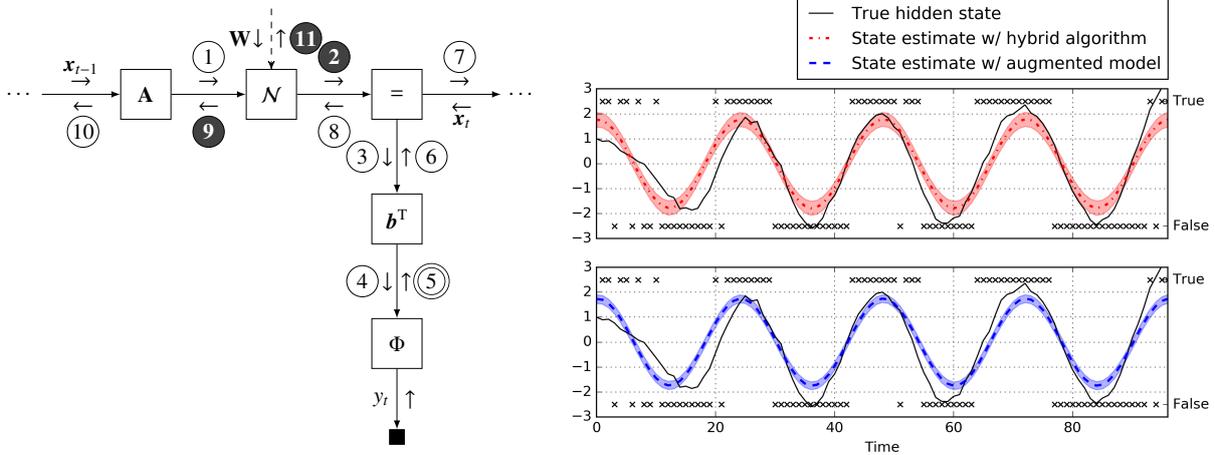

Expectation propagation (EP) \cite{minka_expectation_2001} is an alternative principled approximate Bayesian message passing algorithm that leads to accurate estimates for binary models \cite{kuss_assessing_2005}. For a backward message on the binary node, the EP update is drawn in Fig.~\ref{fig:sigmoid} (right), where the EP message \bwcircled{2} is computed from observation $b$ (which is the incoming message to node $\Phi$ from the right), together with a so-called \emph{cavity} message \circled{1}, which is simply the incoming message to node $\Phi$ from the left side of the FFG. Since EP messages introduce circular message dependencies in the schedule, proper EP-based inference is an iterative algorithm, where multiple iterations of the message passing schedule (hopefully) lead to a stable posterior estimate. For more details on the EP message update for message passing on an FFG, see \cite{cox_robust_nodate}. Using a local EP message at the $\Phi$ node naturally leads to a hybrid VMP-EP schedule, as illustrated in Fig.~\ref{fig:csdo} (left).

We compared the performance of the hybrid VMP-EP algorithm with a more conventional model augmentation technique \cite{albert_bayesian_1993} that allows for full VMP estimation. We inferred the posterior results by performing fifty iterations of both algorithms. The resulting state estimates are shown in Fig.~\ref{fig:csdo} (right). The posterior precisions are \\$\mathcal{W}\left(\matr{W} \,\middle|\, \matr{V}=\begin{pmatrix}31.5 & 0.0 \\ 0.0 & 31.5\end{pmatrix}, \nu=98 \right)$ for the hybrid algorithm and $\mathcal{W}\left(\matr{W} \,\middle|\, \matr{V}=\begin{pmatrix}32.2 & 0.0 \\ 0.0 & 32.2\end{pmatrix}, \nu=98 \right)$ for the augmented model. In order to assess model performance we compared the terminal free energies, which lean $10$ [dB] in favor of the hybrid VMP-EP algorithm.

\FloatBarrier
\subsection{Composite nodes for hierarchical model construction and computational efficiency}
\label{sec:composability}

An important feature of the FFG framework is that a set of neighboring nodes can be grouped together to form a \emph{composite node} that by the rest of the graph is interpreted as a regular single node. Composite nodes hide their internal processing from the rest of the graph, and consequently, inference processes on a graph can proceed as long as each composite node follows the proper message passing communication rules at its interfaces to the rest of the graph.

Composite nodes can be used as modular building blocks in hierarchical model specifications. For instance, a set of nodes can be grouped together as a ``layer''-composite node, and a set of connected ``layer'' nodes can be grouped as a ``layered network''-composite node. This ``layered network''-composite node can now be inserted at any place in any proper FFG, since composite nodes act as regular nodes at their interfaces. In this view, the FFG framework is ``just'' a framework for distributed information processing that specifies how different modules in the network communicate with each other. In a sense, FFGs provide the means to do ``gray-box'' inference in probabilistic models since nodes may hide custom black-box inference procedures.

Since the internal information processing in composite nodes is hidden from the rest of the graph, we can replace parts of the internal message passing computations by other more efficient computations that do not need to be based on message passing. Taking this idea a bit further, it is entirely conceivable to wrap a deep neural network (DNN) from another toolbox into a composite node and use this DNN node as a regular node in our FFG toolbox.

As an example of the use of composite nodes, we reconsider the SSM of Fig.~\ref{fig:lst} (left). Here, the update rule for message $\circled{7}$ requires the inversion of the covariance matrices of the incoming messages $\darkcircled{2}$ and $\circled{6}$ (see \cite{loeliger_introduction_2004}). These inversions might be prohibitively expensive when the dimensionality of the hidden state is large.

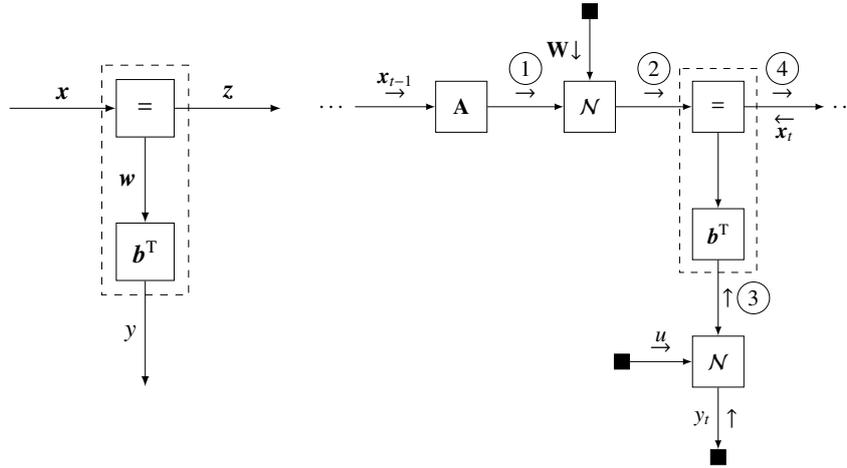
\begin{figure}[h]
    \center
    \begin{subfigure}{.25\textwidth}
        \center
        \resizebox{\textwidth}{!}{\begin{tikzpicture}
    [node distance=20mm,auto,>=stealth']


    \node[box] (eq) {$=$};
    \node[left of=eq] (x) {};
    \node[right of=eq] (z) {};
    \node[box, below of=eq] (b) {$\vect{b}^{\T}$};
    \node[below of=b] (y) {};

    \draw[dashed] (-0.6, 0.6) rectangle (0.6, -2.6);

    \path[line] (x) edge[->] node[anchor=south]{$\vect{x}$} (eq);
    \path[line] (eq) edge[->] node[anchor=south]{$\vect{z}$} (z);
    \path[line] (eq) edge[->] node[anchor=east]{$\vect{w}$} (b);
    \path[line] (b) edge[->] node[anchor=east]{$y$} (y);
\end{tikzpicture}}
    \end{subfigure}
    \begin{subfigure}{.45\textwidth}
        \center
        \resizebox{\textwidth}{!}{\begin{tikzpicture}
    [node distance=20mm,auto,>=stealth']


    \node[box] (A) {$\matr{A}$};
    \node[left of=A] (x_t_min_dots) {$\dots$};
    \node[box, right of=A] (N_x) {$\mathcal{N}$};
    \node[clamped, above of=N_x, node distance=15mm] (W) {};
    \node[box, right of=N_x] (x_eq) {$=$};
    \node[right of=x_eq] (x_t_dots) {$\dots$};

    \node[box, below of=x_eq] (b) {$\vect{b}^{\T}$};
    \node[box, below of=b] (N_y) {$\mathcal{N}$};
    \node[clamped, left of=N_y, node distance=15mm] (u) {};
    \node[clamped, below of=N_y, node distance=15mm] (y_t) {};

    \draw[dashed] (3.4, 0.6) rectangle (4.6, -2.6);

    \path[line] (x_t_min_dots) edge[->] node[anchor=south, yshift=2mm]{$\vect{x}_{t-1}$} node[anchor=south]{$\rightarrow$} (A);
    \path[line] (A) edge[->] (N_x);
    \path[line] (W) edge[->] node[anchor=east, xshift=-2mm]{$\matr{W}$} node[anchor=east]{$\downarrow$} (N_x);
    \path[line] (N_x) edge[->] (x_eq);
    \path[line] (x_eq) edge[->] node[anchor=north, yshift=-2mm]{$\vect{x}_t$} node[anchor=north]{$\leftarrow$} (x_t_dots);

    \path[line] (x_eq) edge[->] (b);
    \path[line] (b) edge[->] (N_y);
    \path[line] (u) edge[->] node[anchor=south, yshift=2mm]{$u$} node[anchor=south]{$\rightarrow$} (N_y);
    \path[line] (N_y) edge[->] node[anchor=east]{$y_t$} node[anchor=west]{$\uparrow$} (y_t);

    \msg{up}{right}{A}{N_x}{0.5}{1};
    \msg{up}{right}{N_x}{x_eq}{0.5}{2};
    \msg{right}{up}{N_y}{b}{0.5}{3};
    \msg{up}{right}{x_eq}{x_t_dots}{0.5}{4};
\end{tikzpicture}}
    \end{subfigure}
    \caption{FFG for the gain-equality composite node (left) and estimation schedule for the linear Gaussian SSM of Sec.~\ref{sec:lgm} with a gain-equality composite node included (right).}
    \label{fig:composite}
\end{figure}

The ``gain-equality''-composite node, as defined in \cite{loeliger_introduction_2004}, avoids the need for inverting large covariance matrices by grouping the equality and observation matrix (vector $\vect{b}$ in our case) into a single node and utilizing the matrix inversion lemma to redefine the message update rule $\circled{4}$ in Fig.~\ref{fig:composite}. As a result, online state estimation in an SSM with large state vectors proceeds more stable and with less computations through the composite node construct. In an FFG, we indicate composite nodes by dashed boxes, see  Fig.~\ref{fig:composite} (left).

ForneyLab provides convenient support to define composite nodes. In order to build this model with ForneyLab, we first define the gain-equality composite node (Fig.~\ref{lst:composite}). This composite node can now be used in the construction of the generative model graph.
\begin{figure}[ht]
\begin{minted}[baselinestretch=1.2,bgcolor=LightGray,fontsize=\footnotesize]{julia}
@composite GainEquality (y, x, z) begin
    @RV w = equal(x, z)
    b = [1.0, 0.0]
    @RV y = dot(b, w)
end
\end{minted}
\caption{Julia code for constructing the gain-equality composite node with ForneyLab. The \texttt{@composite} macro header specifies the new node type (\texttt{GainEquality}) and an ordered tuple of connected variables \texttt{(y, x, z)}. The macro body then defines the (statistical) relations between these (and any internal auxiliary) variables through the standard ForneyLab model definition syntax.}
\label{lst:composite}
\end{figure}

If we do nothing else, then the internal message passing in the composite node is the same as without the composite node definition. However, ForneyLab supports creation of custom update rules inside the composite node. For instance, a custom sum-product rule for message $\circled{4}$ in Fig.~\ref{fig:composite} (right) for the gain-equality node is registered with ForneyLab by the code as shown in Fig.~\ref{lst:rule}.
\begin{figure}[ht]
\begin{minted}[baselinestretch=1.2,bgcolor=LightGray,fontsize=\footnotesize]{julia}
@sumProductRule(:node_type     => GainEquality, # Node type the rule pertains to
                :outbound_type => Message{GaussianMeanVariance}, # Resulting message type from update
                :inbound_types => (Message{Gaussian}, Message{Gaussian}, Void), # Argument message types
                :name          => SPGainEqualityIn2GGV) # Unique rule identifier
\end{minted}
\caption{Julia code for registering a custom gain-equality update rule with ForneyLab. The \texttt{@sumProductRule} macro specifies a sum-product update rule by defining an outbound message type for a specific node-message inputs combination. The rule is given a unique name so that its actual computation can be independently implemented by the inference engine.}
\label{lst:rule}
\end{figure}

In summary, composite nodes provide a very powerful mechanism to build hierarchical models and to customize the internal inference processes in these nodes. Customized rules may make use of convenient re-parameterizations, algebraic tricks or sampling methods that could potentially be implemented by external tools and algorithms. The option to leverage external tools for executing message updates also opens up the possibility to \emph{learn} complex updates rules from the data by means of amortization techniques \cite{stuhlmuller_learning_2013,gershman_amortized_2014}.

\FloatBarrier
\section{Experimental evaluation}
\label{sec:efficient}
In this section we evaluate the usefulness and efficiency of (automatically generated) message passing algorithms for Bayesian inference.
To this end, we consider two common scenarios in Bayesian signal processing: Bayesian parameter estimation in a random walk model (Sec.~\ref{sec:random_walk}) and online learning of the parameters of a linear time-invariant state-space model (Sec.~\ref{sec:co2}). We compare the message passing algorithms as implemented with ForneyLab to MCMC and black box variational (ADVI) methods as implemented in probabilistic programming platforms Stan \cite{carpenter_stan:_2017,kucukelbir_automatic_2015} and Edward \cite{tran_edward:_2016}.

\subsection{Bayesian learning of a random walk model}
\label{sec:random_walk}
Learning the parameters of a random walk model with a latent drift component from noisy observations is a common task in signal processing systems.
Here we consider the task of full Bayesian estimation of all model parameters, and compare the predictive accuracies and running times of multiple algorithms implemented in ForneyLab, Stan and Edward.
Both Monte Carlo and variational algorithms can be viewed as successive approximation methods: the more iterations are performed, the better the list of samples or the variational distribution will approximate the true posterior distribution. Therefore, our goal here is to evaluate the accuracy of multiple algorithm implementations as a function of execution time. Through comparing these performance curves we aim to position ForneyLab in the landscape of automated inference toolboxes in terms of running time versus accuracy.

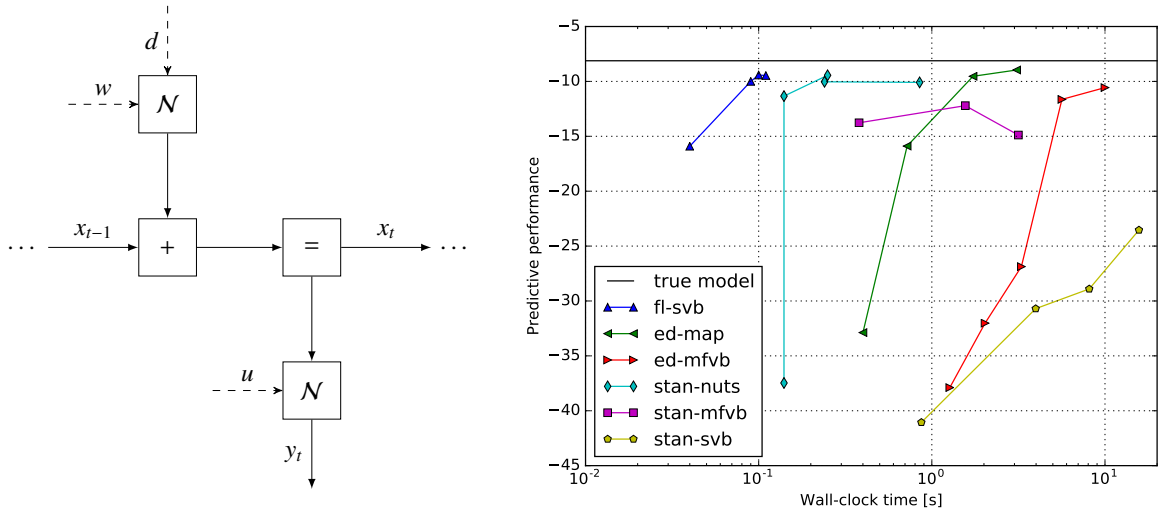
\begin{figure}[h]
    \center
    \begin{subfigure}{.39\textwidth}
        \center
        \resizebox{\textwidth}{!}{\begin{tikzpicture}
    [node distance=20mm,auto,>=stealth']


    \node[box] (add) {$+$};
    \node[left of=add] (x_t_min_dots) {$\dots$};
    \node[box, above of=add] (N_x) {$\mathcal{N}$};
    \node[above of=N_x, node distance=15mm] (d) {};
    \node[left of=N_x, node distance=15mm] (w) {};
    \node[box, right of=add] (x_eq) {$=$};
    \node[right of=x_eq] (x_t_dots) {$\dots$};


    \node[box, below of=x_eq] (N_y) {$\mathcal{N}$};
    \node[left of=N_y, node distance=15mm] (u) {};
    \node[below of=N_y, node distance=15mm] (y_t) {};

    \path[line] (x_t_min_dots) edge[->] node[anchor=south]{$x_{t-1}$} (add);
    \path[line] (N_x) edge[->] (add);
    \path[dashed] (d) edge[->] node[anchor=east]{$d$} (N_x);
    \path[dashed] (w) edge[->] node[anchor=south]{$w$} (N_x);
    \path[line] (add) edge[->] (x_eq);
    \path[line] (x_eq) edge[->] node[anchor=south]{$x_t$} (x_t_dots);

    \path[line] (x_eq) edge[->] (N_y);
    \path[dashed] (u) edge[->] node[anchor=south]{$u$} (N_y);
    \path[line] (N_y) edge[->] node[anchor=east]{$y_t$} (y_t);
\end{tikzpicture}}
    \end{subfigure}
    \begin{subfigure}{.59\textwidth}
        \center
        \resizebox{\textwidth}{!}{\includegraphics{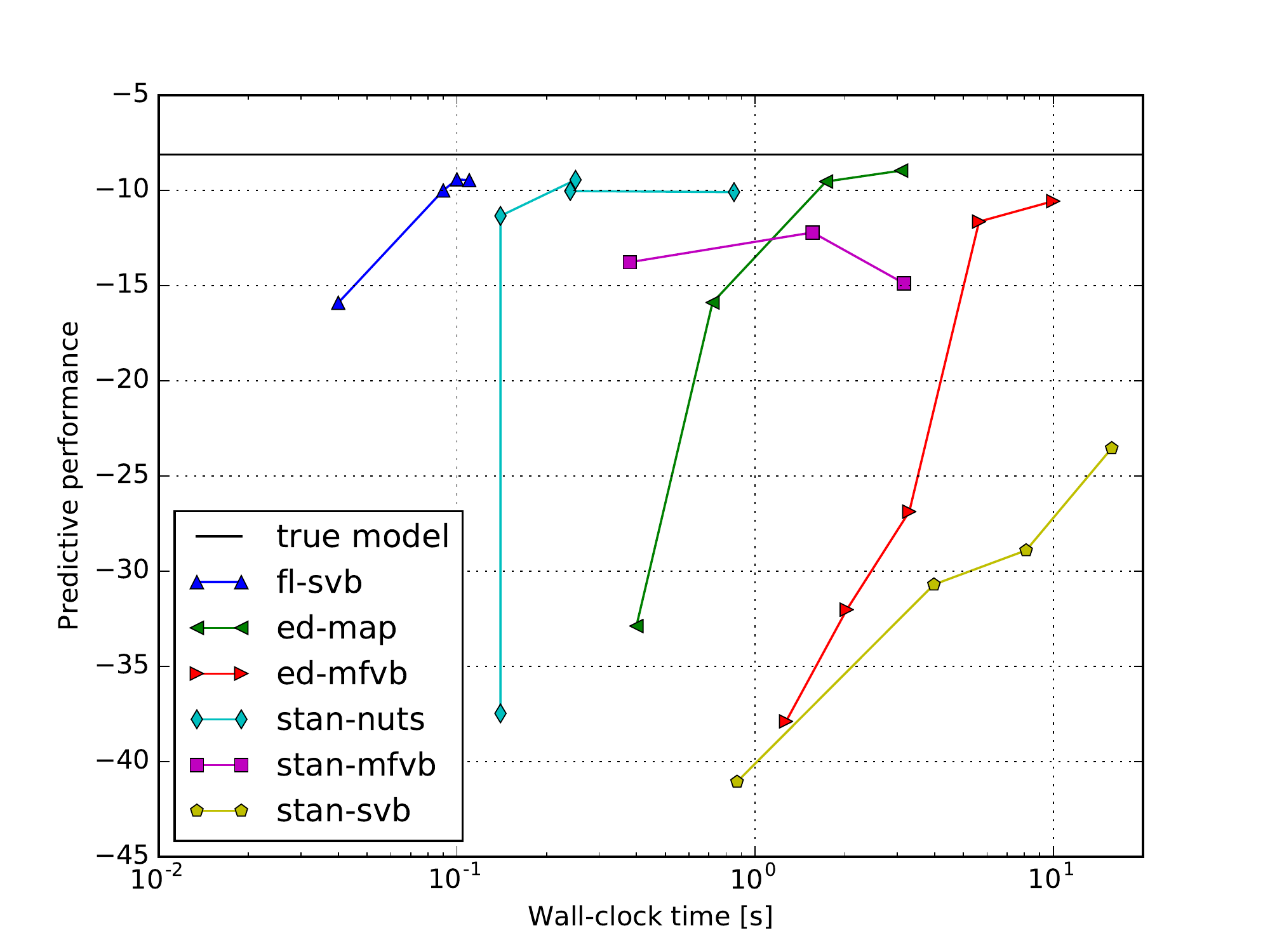}}
    \end{subfigure}
    \caption{FFG representation of the considered random walk model (left) and predictive performance vs. running time for multiple inference methods (right) applied to the random walk model. The markers correspond to runs of the respective inference algorithms for varying iteration counts.} \label{fig:random_walk}
\end{figure}

We consider the following Gaussian random walk model with drift parameter $d$ and Gaussian observation noise:
\begin{subequations}
\label{eq:random-walk}
\begin{align}
    p(x_t \,|\, x_{t-1}, d, w) &= \mathcal{N}\left(x_t \,\middle|\, x_{t-1} + d, w^{-1}\right) \label{eq:rw_transition}\\
    p(y_t \,|\, x_t, u) &= \mathcal{N}\left(y_t \,\middle|\, x_t, u^{-1}\right)\,. \label{eq:rw_observation}
\end{align}
\end{subequations}
The left panel of Fig.~\ref{fig:random_walk} depicts the FFG representation of this model.

The goal is to perform full Bayesian inference of the hidden state sequence as well as the drift and noise parameters. Initial state $x_0$ and drift parameter $d$ are endowed with vague Gaussian priors; the priors on the noise precisions are chosen to be vague Gamma distributions. We perform inference using a variety of inference methods and toolboxes:
\begin{itemize}
    \item No U-Turn Sampling (NUTS, an MCMC method) with Stan (``stan-nuts'');
    \item Mean-field and full-rank ADVI with Stan (``stan-mfvb'' and ``stan-svb'');
    \item MAP inference and mean-field ADVI with Edward (``ed-map'' and ``ed-mfvb'');
    \item Structured VMP with ForneyLab (``fl-svb'').
\end{itemize}
These methods propose different factorizations of the recognition distribution. As a general rule, less factorization assumptions in the recognition distributions is expected to lead to better approximations of the true posterior. On the other hand, a lower degree of factorization also makes it harder to quickly converge to a local minimum in the variational free energy. Full-rank ADVI as performed by Stan assumes no factorization of the recognition distribution. ForneyLab and Edward use the following mean-field and structured factorizations:
\begin{subequations}
\begin{align}
    \text{Mean-field:} & \quad q(x, d, w, u) = q(d)\, q(w)\, q(u) \prod_{t=0}^T q(x_t) \label{eq:rw_q_mf}\\
    \text{Structured:} & \quad q(x, d, w, u) = q(d)\, q(w)\, q(u)\, q(x)\,. \label{eq:rw_q_struct}
\end{align}
\end{subequations}

Inference is performed on a toy data set consisting of $50$ samples drawn from a random walk with drift $\hat{d} = -0.1$, transition precision $\hat{w} = 100$ and observation precision $\hat{u} = 10$.
For performance evaluation, we draw $N=1000$ trajectories $y_{\text{pred}}^{(n)}$ of length $20$ from the true generative process. Then, for each estimation method, we draw $S=1000$ samples from the (approximate) posterior distribution over the parameters. As a performance measure for the inference procedure, we evaluate the average marginal log-likelihood of each sample-trajectory combination, as defined by
\begin{align*}
    Q = \frac{1}{S}\frac{1}{N} \sum_{s=1}^S \sum_{n=1}^N \lognb{p_s \left( y_{\text{pred}}^{(n)} \right) }\,,
\end{align*}
where $p_s$ is the predictive distribution under sampled parameter setting $s$. On average, this metric will favor the approximate posterior with the best predictive performance. Fig.~\ref{fig:random_walk} (right) depicts predictive performance versus running time for the different inference methods and varying iteration counts. All experiments were executed on the same Linux notebook with 16 GB of memory and no GPU acceleration. Roughly speaking, we see that ForneyLab simulations lead to similar or better predictive performance in less (wall-clock) time.

\FloatBarrier
\subsection{Learning a state-space model through streaming variational Bayes}
\label{sec:co2}
\begin{figure}[ht]
    \center
    \begin{subfigure}{.49\textwidth}
        \center
        \resizebox{\textwidth}{!}{\begin{tikzpicture}
    [node distance=20mm,auto,>=stealth']


    \node[box] (A) {$\matr{A}$};
    \node[left of=A] (x_t_min_dots) {$\dots$};
    \node[box, right of=A] (N_x) {$\mathcal{N}$};
    \node[above of=N_x, node distance=15mm] (W) {};
    \node[box, right of=N_x] (x_eq) {$=$};
    \node[right of=x_eq] (x_t_dots) {$\dots$};

    \path[line] (x_t_min_dots) edge[->] node[anchor=south]{$\vect{x}_{t-1}$} (A);
    \path[line] (A) edge[->] (N_x);
    \path[dashed] (W) edge[->] node[anchor=east]{$\matr{W}$} (N_x);
    \path[line] (N_x) edge[->] (x_eq);
    \path[line] (x_eq) edge[->] node[anchor=south]{$\vect{x}_t$} (x_t_dots);

    \node[box, above of=A, xshift=1cm, node distance=2.5cm] (z_add) {$+$};
    \node[box, above of=z_add] (N_z) {$\mathcal{N}$};
    \node[above of=N_z, node distance=15mm] (d) {};
    \node[left of=N_z, node distance=15mm] (g) {};
    \node[left of=z_add, node distance=3cm] (z_t_min_dots) {$\dots$};
    \node[box, right of=z_add] (z_eq) {$=$};
    \node[right of=z_eq, node distance=3cm] (z_t_dots) {$\dots$};


    \path[dashed] (d) edge[->] node[anchor=east]{$d$} (N_z);
    \path[dashed] (g) edge[->] node[anchor=south]{$\gamma$} (N_z);
    \path[line] (N_z) edge[->] (z_add);
    \path[line] (z_t_min_dots) edge[->] node[anchor=south, pos=0.27]{$z_{t-1}$} (z_add);
    \path[line] (z_add) edge[->] (z_eq);
    \path[line] (z_eq) edge[->] node[anchor=south, pos=0.73]{$z_t$} (z_t_dots);

    \node[box, below of=x_eq] (b) {$\vect{b}^{\T}$};
    \coordinate[below of=z_eq, node distance=6.5cm] (z_eq_below);
    \node[box, below of=b] (y_add) {$+$};
    \node[box, below of=y_add] (N_y) {$\mathcal{N}$};
    \node[left of=N_y, node distance=15mm] (u) {};
    \node[below of=N_y, node distance=15mm] (y_t) {};

    \draw[dashed] (3.4, -3.4) rectangle (4.6, -6.6);

    \path[line] (x_eq) edge[->] (b);
    \path[line] (b) edge[->] (y_add);
    \draw (z_eq) -- (z_eq_below) edge[->] (y_add);
    \path[line] (y_add) edge[->] (N_y);
    \path[dashed] (u) edge[->] node[anchor=south]{$u$} (N_y);
    \path[line] (N_y) edge[->] node[anchor=east]{$y_t$} (y_t);
\end{tikzpicture}}
    \end{subfigure}
    \begin{subfigure}{.49\textwidth}
        \center
        \resizebox{\textwidth}{!}{\includegraphics{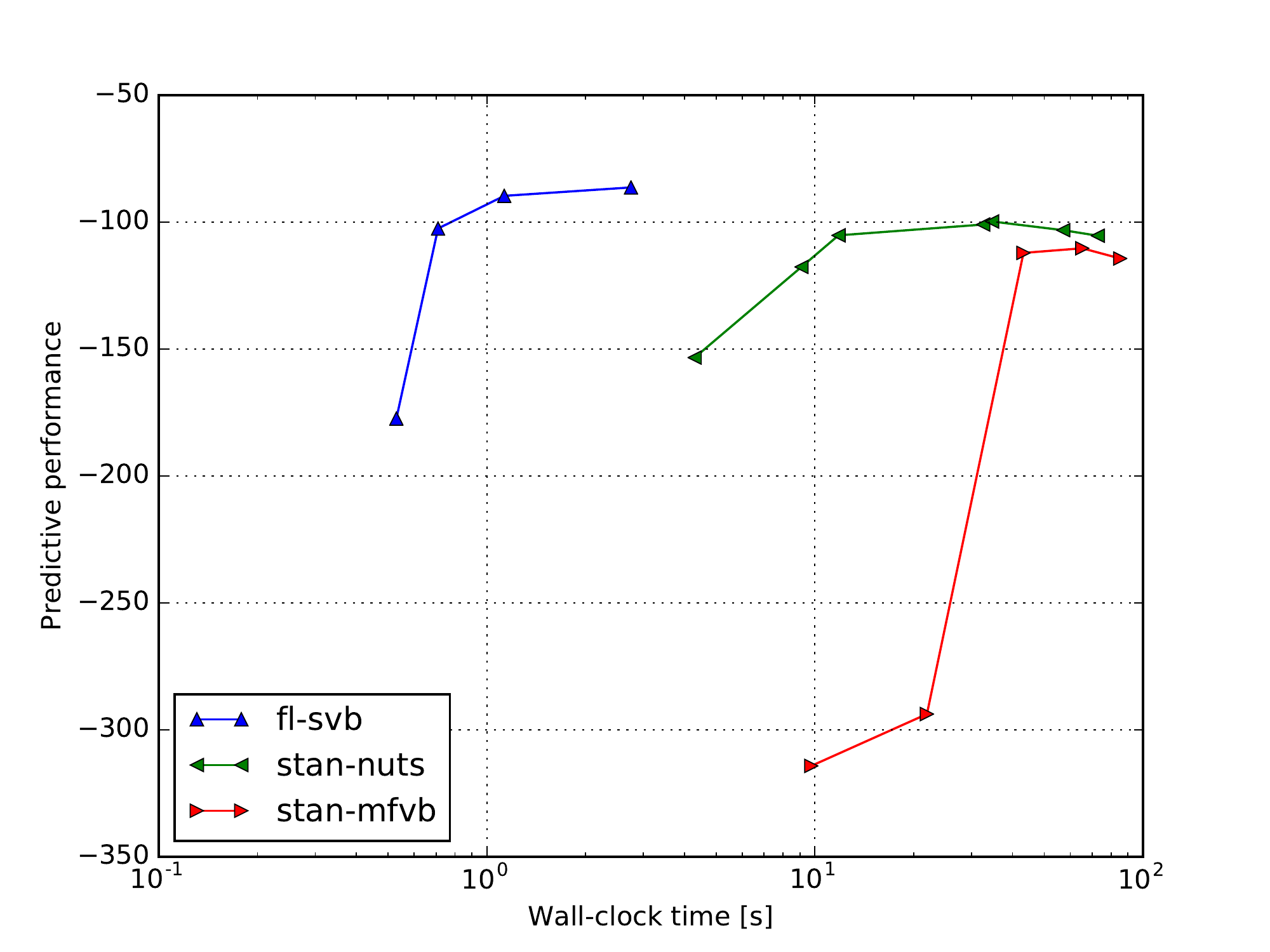}}
        \resizebox{\textwidth}{!}{\includegraphics{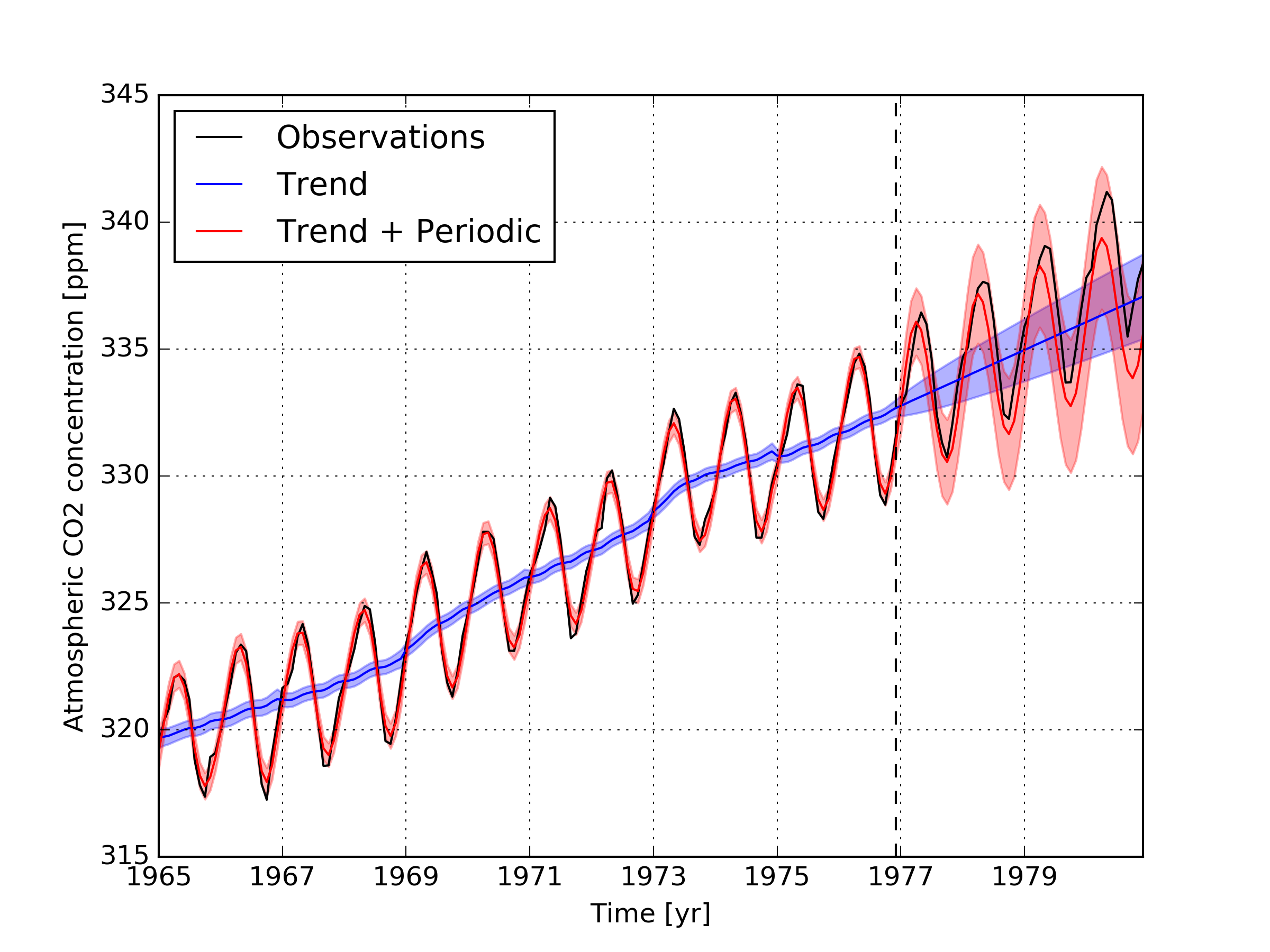}}
    \end{subfigure}
    \caption{Left: FFG representation of the model for CO$_2$ concentration levels. Top right: predictive performance as a function of running time for different inference implementations. Bottom right: CO$_2$ concentration data set and visualization of the inference result obtained by fl-svb. The shaded area corresponds to two standard deviations; the area to the right of the dashed line corresponds to the predictive distribution of the model under the inferred posterior.}
    \label{fig:co2}
\end{figure}
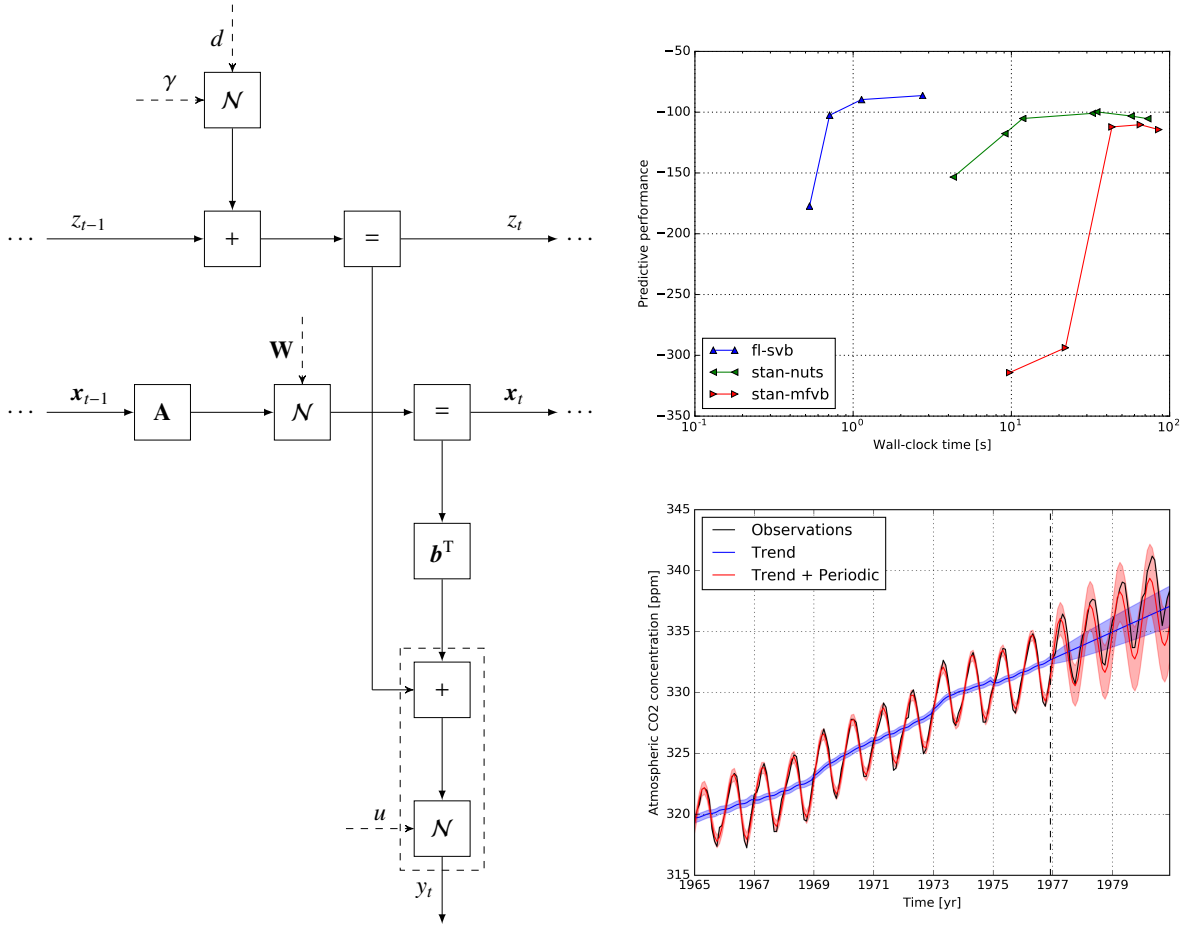

To strain the toolboxes a bit further, in this section we combine the linear Gaussian model of Sec.~\ref{sec:lgm} with the random walk model of Sec.~\ref{sec:random_walk} into a combined SSM, and perform Bayesian inference on a real-world time series. The considered data set is comprised of monthly atmospheric CO$_2$ concentration measurements \cite{hipel_time_1994}, which show a (seasonal) periodic component and a slow upward trend, as shown in Fig.~\ref{fig:co2} (bottom right). The model we consider consists of two components: a Gaussian random walk with drift (to capture the trend) and a SSM for periodic signals (to capture the periodicity). The SSM component models periodicity by applying a rotation matrix $\matr{A}$ in the state transition model, where the latent state $\vect{x}$ represents a phasor. Finally, the observations are modeled by adding the two components under Gaussian observation noise:
\begin{subequations}
\begin{align}
    \text{Trend model:} & \quad p(z_t \,|\, z_{t-1}, d, \gamma) = \mathcal{N} \left(z_t \,\middle|\, z_{t-1} + d, \gamma^{-1}\right) \label{eq:co2_trend_transition}\\
    \text{Periodic model:} & \quad p(\vect{x}_t \,|\, \vect{x}_{t-1}, \matr{W}) = \mathcal{N}\left( \vect{x}_t \,\middle|\, \matr{A}\vect{x}_{t-1} \matr{W}^{-1}\right) \label{eq:co2_periodic_transition}\\
    \text{Observation model:} & \quad p(y_t \,|\, \vect{x}_t, z_t, u) = \mathcal{N} \left( y_t \,\middle|\, \vect{b}^{\T}\vect{x}_t + z_t, u^{-1}\right)\,. \label{eq:co2_observation}
\end{align}
\end{subequations}

Performing full Bayesian inference for the model parameters as well as the hidden state sequences is challenging because of the number of latent variables in the model. A common strategy to keep the computational load limited in such cases is to sequentially perform inference based on mini-batches, using the (approximate) posteriors of the previous step as priors in the next. This approach is known as ``streaming variational inference'' \cite{broderick_streaming_2013}, since it yields an inference algorithm whose computational load scales linearly in terms of data size, thus making it possible to process data in a streaming fashion. In this experiment we apply streaming variational inference with mini-batch size $24$. The initial priors are set to be vague. We use the first six mini-batches for learning and the remaining two mini-batches for evaluating the predictive performance of the fitted model. The predictive performance is defined as the average log-likelihood of the test set under 100 samples of the (approximate) posterior.

We compare the same combinations of toolboxes and inference methods as in Sec.~\ref{sec:random_walk}, with the exception of MAP inference since it cannot be applied in a recursive fashion. Unfortunately, we were not able to obtain a converging algorithm in all cases, even after fixing one or more of the model variables and applying informed initializations. In particular, we were unable to construct converging black box variational inference algorithms in Edward for the given model. The ADVI implementation in Stan does converge, but only in case of mean-field factorization. Fig.~\ref{fig:co2} (top right) contains the results for the methods that converged.

Again, these performance results support the notion that message passing-based inference in factor graphs (as implemented by ForneyLab) is a competitive probabilistic modeling strategy for streaming data applications. In our opinion, when dealing with time-constrained inference and learning problems in dynamical models, message passing-based inference should be a strong candidate inference strategy.

\FloatBarrier
\section{Related work}
\label{sec:related_work}
Interest of the machine learning community in probabilistic programming toolboxes has exploded over the last few years. This rise in popularity is catalyzed by the development of TensorFlow \cite{abadi_tensorflow:_2016}
as a basis for recent toolboxes such as Edward \cite{tran_edward:_2016} and ZhuSuan \cite{shi_zhusuan:_2017}. These toolboxes exploit the analytic expressions of the free energy functional as their optimization objectives. In practice, assumptions about conjugacy or the form of the recognition distribution may limit the scope of workable models. In contrast, sampling-based toolboxes such as PyMC3 \cite{salvatier_probabilistic_2016} (based on Theano \cite{bergstra_theano:_2010}) and Stan \cite{carpenter_stan:_2017} are in general more flexible in their available modeling choices, but pay a price in terms of estimation time.

Modularity in terms of model specification and computation is another important topic in probabilistic programming, because it allows for efficient re-use of pre-specified model and inference primitives as well as for custom extensions upon existing building blocks. Already from the early beginnings, with the development of the BUGS project, the importance of modularity and extensibility of probabilistic programming toolboxes was recognized \cite{lunn_winbugs_2000}. More recent examples of principled approaches to modular model specification are the Bayes Blocks toolbox \cite{harva_bayes_2012}, and the model fragments approach to VMP-based semi-parametric regression \cite{wand_fast_2017}.

Recently, more effort has been directed towards the positioning of probabilistic programming toolboxes in the scientific process. With the development of Edward, the emphasis of the probabilistic programming toolbox shifts from being just a tool for inference, to it being a tool for model criticism as well. This enables probabilistic programming to aid with the full process of iterative model design \cite{blei_build_2014}.

ForneyLab employs the Forney-style factor graph \cite{forney_codes_2001} framework for model representation, and implements VMP as described by \cite{dauwels_variational_2007}. VMP was originally described on Bayesian networks by \cite{winn_variational_2005} and found an implementation in the VIBES framework \cite{bishop_vibes:_2003}. Later implementations of VMP were based on (bipartite) factor graphs \cite{yedidia_constructing_2005}, which found implementations in Infer.NET \cite{minka_infer.net_2018}, Dimple \cite{hershey_accelerating_2012} and Bayes Net \cite{murphy_bayes_2001}.

Bayes Net and Dimple are both based on the MATLAB language, which is practical, but suffers from performance issues for larger models (and MATLAB itself is not free). In contrast, ForneyLab is based on the modern, open source, high-productivity and high-performance language Julia \cite{bezanson_julia:_2017}. Moreover, ForneyLab differs from existing probabilistic programming frameworks because it uses Julia's meta-programming capabilities to produce inference algorithms as executable (Julia) source code. This source code may be readily customized, offering the user precise low-level control over inference execution.

\section{Discussion and conclusions}
\label{sec:discussion}

In this paper we approached Bayesian inference from a message passing perspective, based on a Forney-style factor graph (FFG) description of probabilistic generative models. We painted a broad spectrum of possibilities that arise naturally when adhering to FFGs as a platform for Bayesian inference:
\begin{itemize}
    \item in contrast to ADVI methods, message passing naturally allows for estimation of hybrid models, combining discrete and continuous latent variables (Sec.~\ref{sec:hybrid}).
    \item the modular make-up of the FFG allows for automated derivation of message passing algorithms (Sec.~\ref{sec:lgm});
    \item the automatically derived free energy functional allows for evaluation of the free energy as a model performance measure (Sec.~\ref{sec:lgm});
    \item the modular model representation, together with a principled and automatically derived performance measure, allow for fast iterative search for the best model to fit the data (Sec.~\ref{sec:adaptation});
    \item the message passing formalism allows for combining conceptually distinct inference algorithms under a unified paradigm (Sec.~\ref{sec:combining});
    \item composite nodes allow for implementing custom rules for improved efficiency and flexibility, and allow for hierarchical design in terms of model structure and algorithms (Sec.~\ref{sec:composability});
\end{itemize}
We showed proofs of principle for each of these benefits by implementing examples with ForneyLab, a novel probabilistic programming toolbox for message passing on FFGs.

Tran and Blei summarized four future challenges for message-passing based probabilistic programming languages, in their comment \cite{tran_comment_2017} on Wand \cite{wand_fast_2017}. With ForneyLab, we have addressed (at least in part) these four challenges. The first challenge concerns the specification of local structure by a probabilistic programming language. With ForneyLab, local structure is naturally represented as connections between nodes in an FFG. The user specifies the model through a domain-specific syntax that mimics probabilistic notation, while under the hood the FFG is constructed.

Tran and Blei's second challenge concerned building an extensible inference engine. As mentioned in Sec.~\ref{sec:composability}, in ForneyLab, inference schedules can be naturally extended by including custom (composite) nodes and message updates. However, the extensibility of the inference engine does not end there. The user is free to write her own custom inference engine that takes as input the schedules produced by ForneyLab. This engine might then allow for efficiently combining message updates, or even compile the schedule for implementation on custom hardware.

The third challenge proposes to push modularity even further by introducing hierarchicality in algorithms. We touched upon this idea in Sec.~\ref{sec:composability}, where composite nodes implement custom update rules that may employ any external toolbox or algorithm it can interface with. This allows for building hierarchies of local algorithms. Combinations of inference algorithms were further explored in Sec.~\ref{sec:combining}, where the message passing paradigm allowed for proposing a combined VMP-EP algorithm.

Tran and Blei's fourth challenge related to adapting inference to the computational budget of the user. With message passing, updates can be budgeted based on the information contents of incoming messages. For example, when the entropy of incoming messages is high, the information content of the outbound message might not justify the computational expense. With ForneyLab, budgeting mechanisms may be readily incorporated into the message update rules.

We positioned ForneyLab in the probabilistic toolbox landscape by comparing predictive performance and running time of message passing algorithms as implemented with ForneyLab, with similar inference procedures as implemented with Edward and Stan. For the two SSM problems considered in this paper, ForneyLab exhibited similar or better predictive performance in much less (wall-clock) time. Therefore, we surmise that ForneyLab is a candidate probabilistic modeling framework for automated inference in dynamical systems. These systems are particularly prevalent in the signal processing and control theory communities.

\subsection{Limitations}
The choice of message passing as a platform for probabilistic inference implies some inherent limitations. Message passing relies on the ability of the generative model to factorize into local node functions, which is not always possible. For example, a Gaussian Process model requires a full joint distribution over the modeled (state) variables, disallowing local factorization assumptions on which message passing is based.
A second limitation is due to ForneyLab's requirement for tractable messages. With ForneyLab, messages are represented by a message type together with a finite set of parameters (sufficient statistics). While this representation enables efficient computations, it also limits the flexibility of the posterior distribution. An extension of the message representation might allow for implementation of non-parametric message types such as particle lists.
Finally, ForneyLab does not use parallelization and vectorization out of the box. However, with message passing there is ample opportunity to apply these techniques. More efficient schedulers or engines that account for parallel structure might significantly increase performance.

\subsection{Why Julia?}
Julia is a high-level programming language for technical computing that combines \emph{productivity} with \emph{performance}. The MATLAB-like syntax enables fast and productive prototyping, while the just-in-time (JIT) compiler ensures high performance with execution times close to compiled \emph{C} code \cite{bezanson_julia:_2017}. These properties make Julia an excellent language choice for probabilistic programming, where abstract model and algorithm definitions must be compiled to fast executable inference algorithms.

As already shown in the examples above, ForneyLab takes advantage of Julia's qualities. Firstly, productive sessions with ForneyLab are facilitated through Julia's excellent meta-programming functionalities. The domain-specific syntax for defining the generative model employs macros that convert the abstract model declaration expressions to specific calls to the node constructors that construct the FFG under the hood. Meta-programming is also extensively used on the other end of the modeling pipeline, where high-level schedules are converted to executable Julia code. In essence, ForneyLab is a Julia program that outputs Julia programs. This is advantageous since the complete open source Julia suite (base language and over 1800 registered packages, see \url{https://pkg.julialang.org}) remains seamlessly accessible while working with ForneyLab.

Furthermore, with Julia's multiple dispatch functionality, message computation rules are defined as factor node-specific functions. For varying input message types, a single update rule may yield different outbound message types. Multiple dispatch allows to group separate methods for varying input signatures under a single function name. This greatly improves clarity and effectively implements an innate message lookup-table.

Finally, the open source Julia license makes it a highly accessible language for science and engineering communities across the academic and industrial worlds.

\section*{Acknowledgments}
The authors gratefully acknowledge the contributions made to ForneyLab by Ivan Bocharov and Anouk van Diepen at the Signal Processing Systems (SPS) group at the Eindhoven University of Technology. Furthermore, we thank Ismail Senoz, Tjalling Tjalkens and Wouter van Roosmalen at the SPS group, and Joris Kraak at GN Hearing for stimulating discussions. Finally, we thank the two anonymous reviewers for their valuable comments. This research was made possible by financial support from GN Hearing.

\section*{References}
\bibliographystyle{elsarticle-num}
\bibliography{bibliography}

\begin{thebibliography}{10}
\expandafter\ifx\csname url\endcsname\relax
  \def\url#1{\texttt{#1}}\fi
\expandafter\ifx\csname urlprefix\endcsname\relax\def\urlprefix{URL }\fi
\expandafter\ifx\csname href\endcsname\relax
  \def\href#1#2{#2} \def\path#1{#1}\fi

\bibitem{blei_build_2014}
D.~M. Blei,
  \href{http://www.annualreviews.org/doi/abs/10.1146/annurev-statistics-022513-115657}{Build,
  {Compute}, {Critique}, {Repeat}: {Data} {Analysis} with {Latent} {Variable}
  {Models}}, Annual Review of Statistics and Its Application 1~(1) (2014)
  203--232.
\newblock \href {http://dx.doi.org/10.1146/annurev-statistics-022513-115657}
  {\path{doi:10.1146/annurev-statistics-022513-115657}}.
\newline\urlprefix\url{http://www.annualreviews.org/doi/abs/10.1146/annurev-statistics-022513-115657}

\bibitem{lunn_winbugs_2000}
D.~J. Lunn, A.~Thomas, N.~Best, D.~Spiegelhalter,
  \href{https://link.springer.com/article/10.1023/A:1008929526011}{{WinBUGS} -
  {A} {Bayesian} modelling framework: {Concepts}, structure, and
  extensibility}, Statistics and Computing 10~(4) (2000) 325--337.
\newblock \href {http://dx.doi.org/10.1023/A:1008929526011}
  {\path{doi:10.1023/A:1008929526011}}.
\newline\urlprefix\url{https://link.springer.com/article/10.1023/A:1008929526011}

\bibitem{ranganath_black_2014}
R.~Ranganath, S.~Gerrish, D.~Blei,
  \href{http://proceedings.mlr.press/v33/ranganath14.html}{Black {Box}
  {Variational} {Inference}}, in: {PMLR}, 2014, pp. 814--822.
\newline\urlprefix\url{http://proceedings.mlr.press/v33/ranganath14.html}

\bibitem{carpenter_stan:_2017}
B.~Carpenter, A.~Gelman, M.~D. Hoffman, D.~Lee, B.~Goodrich, M.~Betancourt,
  M.~Brubaker, J.~Guo, P.~Li, A.~Riddell,
  \href{http://www.jstatsoft.org/v76/i01}{Stan: {A} {Probabilistic}
  {Programming} {Language}}, Journal of Statistical Software 76~(1).
\newblock \href {http://dx.doi.org/10.18637/jss.v076.i01}
  {\path{doi:10.18637/jss.v076.i01}}.
\newline\urlprefix\url{http://www.jstatsoft.org/v76/i01}

\bibitem{tran_edward:_2016}
D.~Tran, A.~Kucukelbir, A.~B. Dieng, M.~Rudolph, D.~Liang, D.~M. Blei,
  \href{https://arxiv.org/abs/1610.09787}{Edward: {A} library for probabilistic
  modeling, inference, and criticism}, arXiv preprint arXiv:1610.09787.
\newline\urlprefix\url{https://arxiv.org/abs/1610.09787}

\bibitem{lienart_expectation_2015}
T.~Lienart, Y.~W. Teh, A.~Doucet,
  \href{http://papers.nips.cc/paper/5674-expectation-particle-belief-propagation}{Expectation
  particle belief propagation}, in: Advances in {Neural} {Information}
  {Processing} {Systems}, 2015, pp. 3609--3617.
\newline\urlprefix\url{http://papers.nips.cc/paper/5674-expectation-particle-belief-propagation}

\bibitem{jitkrittum_kernel-based_2015}
W.~Jitkrittum, A.~Gretton, N.~Heess, S.~M.~A. Eslami, B.~Lakshminarayanan,
  D.~Sejdinovic, Z.~Szab{\'o},
  \href{http://www.auai.org/uai2015/proceedings/papers/235.pdf}{Kernel-{Based}
  {Just}-{In}-{Time} {Learning} for {Passing} {Expectation} {Propagation}
  {Messages}}, in: Proceedings of the {Thirty}-{First} {Conference} (2015),
  Amsterdam, Netherlands, 2015.
\newline\urlprefix\url{http://www.auai.org/uai2015/proceedings/papers/235.pdf}

\bibitem{pearl_reverend_1982}
J.~Pearl, \href{http://www.aaai.org/Papers/AAAI/1982/AAAI82-032.pdf}{Reverend
  {Bayes} on {Inference} {Engines}: {A} {Distributed} {Hierarchical}
  {Approach}}, in: Proceedings of the {Second} {AAAI} {Conference} on
  {Artificial} {Intelligence}, {AAAI}'82, AAAI Press, Pittsburgh, Pennsylvania,
  1982, pp. 133--136.
\newline\urlprefix\url{http://www.aaai.org/Papers/AAAI/1982/AAAI82-032.pdf}

\bibitem{winn_variational_2005}
J.~Winn, C.~M. Bishop,
  \href{http://www.jmlr.org/papers/volume6/winn05a/winn05a.pdf}{Variational
  message passing}, Journal of Machine Learning Research 6~(Apr) (2005)
  661--694.
\newline\urlprefix\url{http://www.jmlr.org/papers/volume6/winn05a/winn05a.pdf}

\bibitem{minka_expectation_2001}
T.~P. Minka,
  \href{http://dl.acm.org/citation.cfm?id=2074022.2074067}{Expectation
  {Propagation} for {Approximate} {Bayesian} {Inference}}, in: Proceedings of
  the {Seventeenth} {Conference} on {Uncertainty} in {Artificial}
  {Intelligence}, {UAI}'01, Morgan Kaufmann Publishers Inc., San Francisco, CA,
  USA, 2001, pp. 362--369.
\newline\urlprefix\url{http://dl.acm.org/citation.cfm?id=2074022.2074067}

\bibitem{dauwels_particle_2006}
J.~Dauwels, S.~Korl, H.-a. Loeliger,
  \href{http://ieeexplore.ieee.org/document/4036329}{Particle {Methods} as
  {Message} {Passing}}, IEEE, 2006, pp. 2052--2056.
\newblock \href {http://dx.doi.org/10.1109/ISIT.2006.261910}
  {\path{doi:10.1109/ISIT.2006.261910}}.
\newline\urlprefix\url{http://ieeexplore.ieee.org/document/4036329}

\bibitem{forney_codes_2001}
G.~Forney, \href{https://ieeexplore.ieee.org/abstract/document/910573}{Codes on
  graphs: normal realizations}, IEEE Transactions on Information Theory 47~(2)
  (2001) 520--548.
\newblock \href {http://dx.doi.org/10.1109/18.910573}
  {\path{doi:10.1109/18.910573}}.
\newline\urlprefix\url{https://ieeexplore.ieee.org/abstract/document/910573}

\bibitem{loeliger_introduction_2004}
H.-A. Loeliger, \href{https://ieeexplore.ieee.org/document/1267047}{An
  introduction to factor graphs}, Signal Processing Magazine, IEEE 21~(1)
  (2004) 28--41.
\newline\urlprefix\url{https://ieeexplore.ieee.org/document/1267047}

\bibitem{korl_factor_2005}
S.~Korl, A factor graph approach to signal modelling, system identification and
  filtering, Ph.D. thesis, Swiss Federal Institute of Technology, Zurich
  (2005).

\bibitem{reller_state-space_2012}
C.~Reller, State-{Space} {Methods} in {Statistical} {Signal} {Processing}:
  {New} {Ideas} and {Applications}, Ph.D. thesis, ETH Zurich (2012).

\bibitem{dauwels_variational_2007}
J.~Dauwels, \href{http://ieeexplore.ieee.org/abstract/document/4557602}{On
  {Variational} {Message} {Passing} on {Factor} {Graphs}}, in: {IEEE}
  {International} {Symposium} on {Information} {Theory}, 2007, pp. 2546--2550.
\newblock \href {http://dx.doi.org/10.1109/ISIT.2007.4557602}
  {\path{doi:10.1109/ISIT.2007.4557602}}.
\newline\urlprefix\url{http://ieeexplore.ieee.org/abstract/document/4557602}

\bibitem{attias_variational_1999}
H.~Attias,
  \href{http://papers.nips.cc/paper/1726-a-variational-baysian-framework-for-graphical-models.pdf}{A
  {Variational} {Baysian} {Framework} for {Graphical} {Models}.}, in: {NIPS},
  Vol.~12, 1999.
\newline\urlprefix\url{http://papers.nips.cc/paper/1726-a-variational-baysian-framework-for-graphical-models.pdf}

\bibitem{blei_variational_2017}
D.~M. Blei, A.~Kucukelbir, J.~D. McAuliffe,
  \href{https://www.tandfonline.com/doi/full/10.1080/01621459.2017.1285773}{Variational
  {Inference}: {A} {Review} for {Statisticians}}, Journal of the American
  Statistical Association 112~(518) (2017) 859--877.
\newblock \href {http://dx.doi.org/10.1080/01621459.2017.1285773}
  {\path{doi:10.1080/01621459.2017.1285773}}.
\newline\urlprefix\url{https://www.tandfonline.com/doi/full/10.1080/01621459.2017.1285773}

\bibitem{bezanson_julia:_2017}
J.~Bezanson, A.~Edelman, S.~Karpinski, V.~Shah,
  \href{https://epubs.siam.org/doi/abs/10.1137/141000671}{Julia: {A} {Fresh}
  {Approach} to {Numerical} {Computing}}, SIAM Review 59~(1) (2017) 65--98.
\newblock \href {http://dx.doi.org/10.1137/141000671}
  {\path{doi:10.1137/141000671}}.
\newline\urlprefix\url{https://epubs.siam.org/doi/abs/10.1137/141000671}

\bibitem{loeliger_factor_2007}
H.-A. Loeliger, J.~Dauwels, J.~Hu, S.~Korl, L.~Ping, F.~R. Kschischang, The
  {Factor} {Graph} {Approach} to {Model}-{Based} {Signal} {Processing},
  Proceedings of the IEEE 95~(6) (2007) 1295--1322.
\newblock \href {http://dx.doi.org/10.1109/JPROC.2007.896497}
  {\path{doi:10.1109/JPROC.2007.896497}}.

\bibitem{kucukelbir_automatic_2017}
A.~Kucukelbir, D.~Tran, R.~Ranganath, A.~Gelman, D.~M. Blei,
  \href{http://www.jmlr.org/papers/volume18/16-107/16-107.pdf}{Automatic
  differentiation variational inference}, The Journal of Machine Learning
  Research 18~(1) (2017) 430--474.
\newline\urlprefix\url{http://www.jmlr.org/papers/volume18/16-107/16-107.pdf}

\bibitem{tucker_rebar:_2017}
G.~Tucker, A.~Mnih, C.~J. Maddison, J.~Lawson, J.~Sohl-Dickstein,
  \href{http://papers.nips.cc/paper/6856-rebar-low-variance-unbiased-gradient-estimates-for-discrete-latent-variable-models.pdf}{{REBAR}:
  {Low}-variance, unbiased gradient estimates for discrete latent variable
  models}, in: Advances in {Neural} {Information} {Processing} {Systems}, 2017,
  pp. 2624--2633.
\newline\urlprefix\url{http://papers.nips.cc/paper/6856-rebar-low-variance-unbiased-gradient-estimates-for-discrete-latent-variable-models.pdf}

\bibitem{mathys_uncertainty_2014}
C.~D. Mathys, E.~I. Lomakina, J.~Daunizeau, S.~Iglesias, K.~H. Brodersen, K.~J.
  Friston, K.~E. Stephan,
  \href{http://journal.frontiersin.org/Journal/10.3389/fnhum.2014.00825/full}{Uncertainty
  in perception and the {Hierarchical} {Gaussian} {Filter}}, Frontiers in Human
  Neuroscience 8 (2014) 825.
\newblock \href {http://dx.doi.org/10.3389/fnhum.2014.00825}
  {\path{doi:10.3389/fnhum.2014.00825}}.
\newline\urlprefix\url{http://journal.frontiersin.org/Journal/10.3389/fnhum.2014.00825/full}

\bibitem{bitzer_perceptual_2014}
S.~Bitzer, H.~Park, F.~Blankenburg, S.~J. Kiebel,
  \href{http://journal.frontiersin.org/Journal/10.3389/fnhum.2014.00102/abstract}{Perceptual
  decision making: drift-diffusion model is equivalent to a {Bayesian} model},
  Frontiers in Human Neuroscience 8 (2014) 102.
\newblock \href {http://dx.doi.org/10.3389/fnhum.2014.00102}
  {\path{doi:10.3389/fnhum.2014.00102}}.
\newline\urlprefix\url{http://journal.frontiersin.org/Journal/10.3389/fnhum.2014.00102/abstract}

\bibitem{cox_robust_nodate}
M.~Cox, B.~de~Vries, Robust {Expectation} {Propagation} in {Factor} {Graphs}
  {Involving} {Both} {Continuous} and {Binary} {Variables}, (Accepted for
  publication).

\bibitem{knowles_non-conjugate_2011}
D.~A. Knowles, T.~Minka,
  \href{http://papers.nips.cc/paper/4407-non-conjugate-variational-message-passing-for-multinomial-and-binary-regression.pdf}{Non-conjugate
  {Variational} {Message} {Passing} for {Multinomial} and {Binary}
  {Regression}}, in: J.~Shawe-Taylor, R.~S. Zemel, P.~L. Bartlett, F.~Pereira,
  K.~Q. Weinberger (Eds.), Advances in {Neural} {Information} {Processing}
  {Systems} 24, Curran Associates, Inc., 2011, pp. 1701--1709.
\newline\urlprefix\url{http://papers.nips.cc/paper/4407-non-conjugate-variational-message-passing-for-multinomial-and-binary-regression.pdf}

\bibitem{nolan_accurate_2017}
T.~H. Nolan, M.~P. Wand,
  \href{http://onlinelibrary.wiley.com/doi/10.1002/sta4.139/abstract}{Accurate
  logistic variational message passing: algebraic and numerical details}, Stat
  6~(1) (2017) 102--112.
\newblock \href {http://dx.doi.org/10.1002/sta4.139}
  {\path{doi:10.1002/sta4.139}}.
\newline\urlprefix\url{http://onlinelibrary.wiley.com/doi/10.1002/sta4.139/abstract}

\bibitem{albert_bayesian_1993}
J.~H. Albert, S.~Chib,
  \href{https://www.tandfonline.com/doi/abs/10.1080/01621459.1993.10476321}{Bayesian
  {Analysis} of {Binary} and {Polychotomous} {Response} {Data}}, Journal of the
  American Statistical Association 88~(422) (1993) 669--679.
\newblock \href {http://dx.doi.org/10.1080/01621459.1993.10476321}
  {\path{doi:10.1080/01621459.1993.10476321}}.
\newline\urlprefix\url{https://www.tandfonline.com/doi/abs/10.1080/01621459.1993.10476321}

\bibitem{kuss_assessing_2005}
M.~Kuss, C.~E. Rasmussen,
  \href{http://www.jmlr.org/papers/volume6/kuss05a/kuss05a.pdf}{Assessing
  approximate inference for binary {Gaussian} process classification}, The
  Journal of Machine Learning Research 6 (2005) 1679--1704.
\newline\urlprefix\url{http://www.jmlr.org/papers/volume6/kuss05a/kuss05a.pdf}

\bibitem{stuhlmuller_learning_2013}
A.~Stuhlm{\"u}ller, J.~Taylor, N.~Goodman,
  \href{http://papers.nips.cc/paper/4966-learning-stochastic-inverses.pdf}{Learning
  stochastic inverses}, in: Advances in neural information processing systems,
  2013, pp. 3048--3056.
\newline\urlprefix\url{http://papers.nips.cc/paper/4966-learning-stochastic-inverses.pdf}

\bibitem{gershman_amortized_2014}
S.~Gershman, N.~Goodman,
  \href{http://escholarship.org/uc/item/34j1h7k5}{Amortized {Inference} in
  {Probabilistic} {Reasoning}}, Proceedings of the Cognitive Science Society
  36~(36).
\newline\urlprefix\url{http://escholarship.org/uc/item/34j1h7k5}

\bibitem{kucukelbir_automatic_2015}
A.~Kucukelbir, R.~Ranganath, A.~Gelman, D.~Blei,
  \href{http://papers.nips.cc/paper/5758-automatic-variational-inference-in-stan.pdf}{Automatic
  {Variational} {Inference} in {Stan}}, in: C.~Cortes, N.~D. Lawrence, D.~D.
  Lee, M.~Sugiyama, R.~Garnett (Eds.), Advances in {Neural} {Information}
  {Processing} {Systems} 28, Curran Associates, Inc., 2015, pp. 568--576.
\newline\urlprefix\url{http://papers.nips.cc/paper/5758-automatic-variational-inference-in-stan.pdf}

\bibitem{hipel_time_1994}
K.~W. Hipel, A.~I. McLeod, Time series modelling of water resources and
  environmental systems, Elsevier, 1994.

\bibitem{broderick_streaming_2013}
T.~Broderick, N.~Boyd, A.~Wibisono, A.~C. Wilson, M.~I. Jordan,
  \href{http://papers.nips.cc/paper/4980-streaming-variational-bayes}{Streaming
  variational bayes}, in: Advances in {Neural} {Information} {Processing}
  {Systems}, 2013, pp. 1727--1735.
\newline\urlprefix\url{http://papers.nips.cc/paper/4980-streaming-variational-bayes}

\bibitem{abadi_tensorflow:_2016}
M.~Abadi, P.~Barham, J.~Chen, Z.~Chen, A.~Davis, J.~Dean, M.~Devin,
  S.~Ghemawat, G.~Irving, M.~Isard,
  \href{https://www.usenix.org/system/files/conference/osdi16/osdi16-abadi.pdf}{{TensorFlow}:
  a system for large-scale machine learning}, in: Proceedings of the 12th
  {USENIX} conference on {Operating} {Systems} {Design} and {Implementation},
  USENIX Association, 2016, pp. 265--283.
\newline\urlprefix\url{https://www.usenix.org/system/files/conference/osdi16/osdi16-abadi.pdf}

\bibitem{shi_zhusuan:_2017}
J.~Shi, J.~Chen, J.~Zhu, S.~Sun, Y.~Luo, Y.~Gu, Y.~Zhou,
  \href{http://arxiv.org/abs/1709.05870}{{ZhuSuan}: {A} {Library} for
  {Bayesian} {Deep} {Learning}}, arXiv:1709.05870 [cs, stat]ArXiv: 1709.05870.
\newline\urlprefix\url{http://arxiv.org/abs/1709.05870}

\bibitem{salvatier_probabilistic_2016}
J.~Salvatier, T.~V. Wiecki, C.~Fonnesbeck,
  \href{https://peerj.com/articles/cs-55}{Probabilistic programming in {Python}
  using {PyMC}3}, PeerJ Computer Science 2 (2016) e55.
\newblock \href {http://dx.doi.org/10.7717/peerj-cs.55}
  {\path{doi:10.7717/peerj-cs.55}}.
\newline\urlprefix\url{https://peerj.com/articles/cs-55}

\bibitem{bergstra_theano:_2010}
J.~Bergstra, O.~Breuleux, F.~Bastien, P.~Lamblin, R.~Pascanu, G.~Desjardins,
  J.~Turian, D.~Warde-Farley, Y.~Bengio,
  \href{http://conference.scipy.org/proceedings/scipy2010/pdfs/bergstra.pdf}{Theano:
  {A} {CPU} and {GPU} math compiler in {Python}}, in: Proc. 9th {Python} in
  {Science} {Conf}, Vol.~1, 2010.
\newline\urlprefix\url{http://conference.scipy.org/proceedings/scipy2010/pdfs/bergstra.pdf}

\bibitem{harva_bayes_2012}
M.~Harva, T.~Raiko, A.~Honkela, H.~Valpola, J.~Karhunen,
  \href{https://arxiv.org/pdf/1207.1380}{Bayes {Blocks}: {An} implementation of
  the variational {Bayesian} building blocks framework}, arXiv preprint
  arXiv:1207.1380.
\newline\urlprefix\url{https://arxiv.org/pdf/1207.1380}

\bibitem{wand_fast_2017}
M.~P. Wand,
  \href{https://www.tandfonline.com/doi/full/10.1080/01621459.2016.1197833}{Fast
  {Approximate} {Inference} for {Arbitrarily} {Large} {Semiparametric}
  {Regression} {Models} via {Message} {Passing}}, Journal of the American
  Statistical Association 112~(517) (2017) 137--168.
\newblock \href {http://dx.doi.org/10.1080/01621459.2016.1197833}
  {\path{doi:10.1080/01621459.2016.1197833}}.
\newline\urlprefix\url{https://www.tandfonline.com/doi/full/10.1080/01621459.2016.1197833}

\bibitem{bishop_vibes:_2003}
C.~M. Bishop, D.~Spiegelhalter, J.~Winn,
  \href{http://papers.nips.cc/paper/2172-vibes-a-variational-inference-engine-for-bayesian-networks.pdf}{{VIBES}:
  {A} variational inference engine for {Bayesian} networks}, in: Advances in
  neural information processing systems, 2003, pp. 793--800.
\newline\urlprefix\url{http://papers.nips.cc/paper/2172-vibes-a-variational-inference-engine-for-bayesian-networks.pdf}

\bibitem{yedidia_constructing_2005}
J.~S. Yedidia, W.~Freeman, Y.~Weiss,
  \href{http://ieeexplore.ieee.org/abstract/document/1459044}{Constructing
  free-energy approximations and generalized belief propagation algorithms},
  IEEE Transactions on Information Theory 51~(7) (2005) 2282--2312.
\newblock \href {http://dx.doi.org/10.1109/TIT.2005.850085}
  {\path{doi:10.1109/TIT.2005.850085}}.
\newline\urlprefix\url{http://ieeexplore.ieee.org/abstract/document/1459044}

\bibitem{minka_infer.net_2018}
T.~Minka, J.~Winn, J.~Guiver, Y.~Zaykov, D.~Fabian, J.~Bronskill,
  \href{http://research.microsoft.com/infernet}{Infer.{NET} 2.7}, microsoft
  Research Cambridge (2018).
\newline\urlprefix\url{http://research.microsoft.com/infernet}

\bibitem{hershey_accelerating_2012}
S.~Hershey, J.~Bernstein, B.~Bradley, A.~Schweitzer, N.~Stein, T.~Weber,
  B.~Vigoda, \href{http://arxiv.org/abs/1212.2991}{Accelerating {Inference}:
  towards a full {Language}, {Compiler} and {Hardware} stack}, arXiv:1212.2991
  [cs, stat].
\newline\urlprefix\url{http://arxiv.org/abs/1212.2991}

\bibitem{murphy_bayes_2001}
K.~Murphy,
  \href{http://interfacesymposia.org/I01/I2001Proceedings/KMurphy/KMurphy.pdf}{The
  bayes net toolbox for matlab}, Computing science and statistics 33~(2) (2001)
  1024--1034.
\newline\urlprefix\url{http://interfacesymposia.org/I01/I2001Proceedings/KMurphy/KMurphy.pdf}

\bibitem{tran_comment_2017}
D.~Tran, D.~M. Blei,
  \href{https://doi.org/10.1080/01621459.2016.1270044}{Comment}, Journal of the
  American Statistical Association 112~(517) (2017) 156--158.
\newblock \href {http://dx.doi.org/10.1080/01621459.2016.1270044}
  {\path{doi:10.1080/01621459.2016.1270044}}.
\newline\urlprefix\url{https://doi.org/10.1080/01621459.2016.1270044}

\end{thebibliography}

\end{document}